\documentclass[twocolumn]{article} 
\usepackage{times}  
\usepackage{helvet}
\usepackage{courier} 
\usepackage{mathtools}
\usepackage[hyphens]{url} 
\usepackage{graphicx}
\usepackage{natbib} 
\usepackage{caption} 
\DeclareCaptionStyle{ruled}{labelfont=normalfont,labelsep=colon,strut=off} 
\usepackage{algorithm}
\usepackage{algorithmic}
\usepackage{newfloat}
\usepackage{listings}
\floatstyle{ruled}
\newfloat{listing}{tb}{lst}{}
\floatname{listing}{Listing}

\usepackage[symbol]{footmisc}

\renewcommand{\thefootnote}{\fnsymbol{footnote}}
\title{Neuro-Symbolic Hierarchical Rule Induction}
\author{
Claire Glanois \footnotemark[2] \quad
Xuening Feng \footnotemark[2] \quad
Zhaohui Jiang \footnotemark[2] \\
Paul Weng \footnotemark[2] \quad
Matthieu Zimmer \footnotemark[2] \quad
Dong Li \footnotemark[3] \quad
Wulong Liu \footnotemark[3]
}

\date{}
\usepackage{amsfonts}
\usepackage{amsmath}
\usepackage{amsthm}
\usepackage{multirow}
\usepackage{appendix}
\usepackage{xcolor}
\usepackage{booktabs}
\usepackage{bm}
\usepackage{etoolbox}
\newtoggle{final}
\toggletrue{final}
\newcommand{\Constants}{\mathcal C}
\newcommand{\Predicates}{\mathcal P}
\newcommand{\PR}{\mathcal R}
\newcommand{\shortminus}{\scalebox{0.75}[1.0]{\( - \)}}
\newcommand{\ours}{HRI } % name of our model/method

\definecolor{dkgreen}{rgb}{0,0.6,0}

\newcommand{\pw}[1]{\iftoggle{final}{#1}{{\color{blue} #1}}}
\newcommand{\TODO}[1]{\iftoggle{final}{}{{\color{red} #1}}}
\newcommand{\cg}[1]{\iftoggle{final}{#1}{{\color{dkgreen} #1}}}
\newcommand{\cgnew}[1]{\iftoggle{final}{#1}{{\color{purple} #1}}}
\newcommand{\xf}[1]{\iftoggle{final}{#1}{{\color{teal} #1}}}
\newcommand{\zj}[1]{\iftoggle{final}{#1}{{\color{orange} #1}}}
\newcommand{\mz}[1]{\iftoggle{final}{#1}{{\color{gray} #1}}}
\newtheorem{theorem}{Theorem}[section]
\newtheorem{lemma}[theorem]{Lemma}
\newtheorem{corollary}[theorem]{Corollary}
\newtheorem{definition}[theorem]{Definition}

\begin{document}

\maketitle

\footnotetext[2]{Shanghai Jiao Tong University}
\footnotetext[3]{Huawei Noah's Ark Lab}

%If you would like to switch back to Arabic numbering, you can do it by
\renewcommand*{\thefootnote}{\arabic{footnote}}
%amd reset counter
\setcounter{footnote}{0}

\begin{abstract}

We propose an efficient interpretable neuro-symbolic model to solve Inductive Logic Programming (ILP) problems.
In this model, which is built from a set of meta-rules organised in a hierarchical structure, first-order rules are invented by learning embeddings to match facts and body predicates of a meta-rule. 
To instantiate \cg{it}, we specifically design an expressive set of generic meta-rules, and demonstrate they generate a consequent fragment of Horn clauses. During training, we inject a controlled \pw{Gumbel} noise to avoid local optima and employ interpretability-regularization term to further guide the convergence to interpretable rules.
%and guide the convergence to interpretable rules, which are further favored with an interpretability-regularization term. 
We empirically validate our model on various tasks (ILP, visual genome, reinforcement learning) against several state-of-the-art methods.
\end{abstract}

%%%%%%%%%%%%%%%%%%%%%%%%%%%%%%%%%%%%%%%%%%%%%%%%%%
\section{Introduction}\label{sec:intro}
%%%%%%%%%%%%%%%%%%%%%%%%%%%%%%%%%%%%%%%%%%%%%%%%%%

Research on neuro-symbolic approaches \citep{santoro_simple_2017,donadello_logic_2017,manhaeve_neural_2019,dai_bridging_2019,garcez2020neurosymbolic} has become very active recently and has been fueled by the successes of deep learning \citep{alexnet} and the recognition of its limitations \citep{marcus_dl_2018}. 
At a high level, these approaches aim at combining the strengths of deep learning (e.g., high expressivity and differentiable learning) and symbolic methods (e.g., interpretability and generalizability) while addressing their respective limitations (e.g., brittleness for deep learning and scalability for symbolic methods).

In this paper, we specifically focus on inductive logic programming (ILP) tasks \citep{cropper_inductive_2020}.
The goal in ILP is to find a first-order logic (FOL) program that explains positive and negative examples given some background knowledge also expressed in FOL.
In contrast to classic ILP methods, which are based on combinatorial search over the space of logic programs, neuro-symbolic methods (see Section~\ref{sec:related}) usually work on a continuous relaxation of this \cg{process}.

For tackling ILP problems, we propose a new neuro-symbolic hierarchical model that is \cg{built upon a} set of meta-rules, denoted \ours.
\cg{For a more compact yet expressive representation}, we introduce the notion of \textit{proto-rule}, which \cg{encompasses} multiple meta-rules.
%In order to compactly express meta-rules, 
The valuations of predicates are computed via a soft unification between proto-rules and predicates using learnable embeddings.
Like most ILP methods, our model can provide an interpretable solution and, being independent of the number of objects,  \cg{manifests some \emph{combinatorial generalisation} skills}\footnote{Trained on smaller instances, it can generalize to larger ones.}.
In contrast to most other approaches, \ours is also independent of the number of predicates \cg{---e.g.} this number may vary between training and testing.
Our model can also allow recursive definitions, if needed.
Since it is based on embeddings, it is able to generalize and is particularly suitable for multi-task ILP problems. 
%, \cg{and rich semantic environments}.
% This is said in the next sentence.
%NB. benefit from embeddings if many initial predicates, and not orthogonal...
\cg{%Our model would thereupon be particularly adapted to a context of \textit{continual learning}.
% It's better not to mention continual learning, because we don't do any experiments on it. We may mention that in the conclusion maybe.
Moreover, \pw{any} semantic or visual priors on concepts can be leveraged, to initialize the predicate embeddings.
%\footnote{For instance, in a context of autonomous driving, the classes (unary predicates) may be bootstrapped both from words from our vocabulary (e.g., "pedestrian", "bike", etc.), or from the bounding boxes embeddings (e.g., extracted from the RCNN).}
% This will be explained in the multi-task experiments.
}

The design of proto-rules, crucially restricting the hypothesis space, embodies a well-known trade-off between efficiency and expressivity. 
Relying on minimal sets of proto-rules for rule induction models has been shown to improve both learning time and predictive accuracies \citep{cropper_logical_2014,fonseca_avoiding_2004}.
For our model to be both adaptive and efficient, we initially designed an expressive and minimal set $\PR_0$ of proto-rules.
%more precisely, we conceived an extended definition of meta-rule, which do not constraint the arity of the body predicates\footnote{Only the arity of the head predicate is determined as the head predicate is fixed in our implementation of meta-rules}.
%We refer to Appendix~\ref{appendix:meta_rules} for a more formal definition.
While most neuro-symbolic approaches do not formally discuss the expressivity of their models, we provide a theoretical analysis \cg{to characterize the} expressivity of $\PR_0$.
Moreover, in contrast to ILP work based on combinatorial search \citep{cropper_logical_2020}, we found that a certain redundanc\mz{y} in the proto-rules may experimentally help learning in neuro-symbolic approaches.
Therefore, as a replacement of $\PR_0$, we propose an extended set $\PR_*$ of proto-rules. % that we use in all our experiments.

We validate our model using proto-rules in $\PR_*$ on classic ILP tasks.
\cg{Despite} using the same set of proto-rules, our model is competitive with other approaches that may require specifying different meta-rules for different tasks to work\pw{, which, \cg{unsatisfactorily}, requires an a priori knowledge of the solution}.
In order to exploit the embeddings of our model and demonstrate its scalability, we \cg{distinctly} evaluate our approach on \pw{a} large domain,
%WN18 \citep{antoine_bordes_translating_2013}, a subset of WordNet and 
GQA \citep{Hudson_Manning_2019} extracted from Visual Genome \citep{krishna_visual_2017}).
Our model outperforms other methods such as NLIL \citep{yang_learn_2019} on those tasks. 
We also empirically validate all our design choices.
% Examples of close predicates in VGA: 
% on vs. over and 
% hanging from vs. attached to

\paragraph{Contributions}
Our contributions can be summarized as follows:
(1) \cg{Hierarchical model for embedding-based rule induction} (see Section~\ref{sec:model}),
%Hierarchical model of embedding-based meta-rules
(2) Expressive set of generic proto-rules and theoretical analysis (see Section~\ref{sec:set}),
(3) \cg{interpretability-oriented} training method (see Section~\ref{sec:training}),
(4) Empirical validation on various domains (see Section~\ref{sec:experiments}).

%%%%%%%%%%%%%%%%%%%%%%%%%%%%%%%%%%%%%%%%%%%%%%%%%%
\section{Related Work}\label{sec:related}
%%%%%%%%%%%%%%%%%%%%%%%%%%%%%%%%%%%%%%%%%%%%%%%%%%

\paragraph{Classic ILP methods} 
Classic symbolic ILP methods \citep{Quinlan90learninglogical,Muggleton2009InverseEA,cropper_logical_2020}
% FOIL: top-down, Progol: top-down (but bottom-up init), bottom-up, meta-level
aim to learn logic rule\mz{s} from examples by a direct search in the space of logic programs.
%strategy (e.g., Metagol \citep{Muggleton_meta-interpretivelearning, cropper_logical_2020}), some turn to a bottom-up approach (e.g., Progol, \citep{Muggleton2009InverseEA}) extracting clauses from examples and aiming to generalize them.
To scale to large knowledge graphs,  recent work \citep{galarraga_amie_2013, omran_scalable_2018} \pw{learns predicate and entity} embeddings to guide and prune the search.
These \pw{methods} typically rely on carefully hand-designed and task-specific templates in order to narrow down the hypothesis space.
\pw{Their drawback is their difficulties with noisy data.}

\paragraph{Differentiable ILP methods} %These \pw{approaches} are \pw{based} on a continuous relaxation of the logical reasoning process, \pw{therefore} the parameters can be trained \pw{with} gradient descent.
%The ILP task is commonly framed as a binary classification problem.
\pw{By framing an ILP task as a classification problem, these approaches based on a continuous relaxation of the logical reasoning process can learn via gradient descent.}
\cg{The learnable weights may be assigned to \pw{rules} \citep{evans_learning_2017}---although combinatorially less attractive---or to the membership of predicates in a clause \citep{payani_learning_2019, zimmer2021differentiable}.}
%The latter idea was also considered in another work \citep{payani_learning_2019}, but their language bias was insufficient.^
%we don’t have to generate all the candidate rules. Instead,
However, as ILP solvers, these models are hard to scale and have a constrained implementation, e.g., %small maximal arity,
% better not to mention that, our model only uses binary predicates
task-specific template-based variable assignments or limited number of predicates and objects.
In another direction, NLM \citep{dong_neural_2019} learns rules as shallow MLPs;
by doing so, it crucially loses in interpretability.
%\cg{This observation motivated \citep{zimmer2021differentiable} to replace the neural modules in NLM by interpretable Boolean modules, while keeping the hierarchical structure present in NLM.}
% Not important for our current paper
In contrast to these \pw{methods}, our model \ours ascribes embeddings to predicates, and the membership weights derive from a similarity score, \pw{which may} 
be beneficial in rich and \pw{multi-task} learning domains.

%(1) arranging the rules in a hierarchical structure where the facts are at the first layer and the target predicate is at the last layer.

\paragraph{Multi-hop reasoning methods}
\cg{In multi-hop reasoning methods, developed around knowledge base (KB) completion tasks, predicate invention is understood as finding a relational path \citep{yang_differentiable_2017} \pw{or a combination of them} \citep{yang_learn_2019} on the underlying knowledge graph;
this path amounts to a chain-like first-order rule.
%for instance, some authors have proposed to use differentiable attention operators to guide this navigation \citep{yang_differentiable_2017}.
%However, the chain-like rules are limited in their expressivity and these paths are query-specific, instead of aiming to learn generic rules.
%These drawbacks have been partially tackled in \citep{yang_learn_2019}.
}%
\pw{However, although computationally efficient, the restricted expressiveness of these methods limit their performances.}

% Highly efficient in in variable binding and rule evaluation, compared to the template-based counterparts??
%Neural LP learns a distribution over logical rules, instead of an approximation to a particular rule.
% learning probabilistic FOL rules from KB,

\paragraph{Embedding-based models}
\cg{In the context of KB completion or rule mining notably, \pw{many} previous studies learn embeddings of both binary predicates and entities.
Entities are typically attached to low-dimensional vectors, while relations are understood as bilinear or linear operators applied on entities %, such as translation \citep{antoine_bordes_translating_2013}
%TransE
%or rotation operators \citep{Lin15learningentity}, 
% TransR
possibly involving some non-linear operators %\citep{Socher_reasoningwith}.
% NTN
}

\pw{\citep{antoine_bordes_translating_2013,Lin15learningentity,Socher_reasoningwith,m_nickel_three-way_2011,guu_traversing_2015}.
These methods are either limited in their reasoning power or suffer from similar problems as multi-hop reasoning.}

\pw{Our work is closely related to that of \citet{campero2018logical} whose representation model} is inspired by NTP \citep{rocktaschel_end--end_2017}, attaching vector embeddings to predicates and rules. 
%However, the latter opts for a backward chaining method, wheareas the former favors a forward chaining methods to avoid tracking down the proof tree.
%Another difference lies in the choice of the similarity metric, since \citep{rocktaschel_end--end_2017} rely on radial basis function kernel, whereas  \citep{campero2018logical} use cosine similarity.
\pw{Their} major drawback, like most \pw{classic or differentiable} ILP methods, is \pw{the need for a} careful\pw{ly} hand-design\pw{ed} template set for each ILP task.
In contrast, we extend their model to a hierarchical structure with an expressive set of meta-rules, which can be used for \pw{various} tasks.
We further demonstrate our approach in the multi-task setting.

%%%%%%%%%%%%%%%%%%%%%%%%%%%%%%%%%%%%%%%%%%%%%%%%%%
\section{Background}\label{sec:background}
%%%%%%%%%%%%%%%%%%%%%%%%%%%%%%%%%%%%%%%%%%%%%%%%%%

We first define first-order logic and inductive logic programming (ILP).
Then, we recall some approaches for \textit{predicate invention}, an essential aspect of ILP. %the model proposed by \citet{campero2018logical}.

\paragraph{Notations} 
Sets are denoted in calligraphic font (e.g., $\Constants$).
Constants (resp. variables) are denoted in lowercase (resp. uppercase).
Predicates have the first letter of their names capitalized (e.g., $P$ or $Even$), their corresponding atoms are denoted sans serif (e.g., $\mathsf P$), while predicate variables are denoted in roman font (e.g., $\mathrm P$).
Integer hyperparameters are denoted $n$ with a subscript (e.g., $n_L$).

\subsection{Inductive Logic Programming}

\textit{First-order logic} (FOL) is a formal language that can be defined with a set of constants $\Constants$, a set of predicates $\Predicates$, a set of functions, and a set of variables.
\textit{Constants} correspond to the objects of discourse, 
$n$-ary \textit{predicates} can be seen as mappings from $\Constants^n$ to the Boolean set $\mathbb B = \{\pw{\mathtt{True}}, \pw{\mathtt{False}}\}$, 
$n$-ary \textit{functions} are mappings from $\Constants^n$ to the set of constants $\Constants$, and
\textit{variables} correspond to unspecified constants.
%Unary predicates (e.g., $Even(X)$ is true if $X$ is an even number) provide properties satisfied by objects and binary predicates (e.g., $Succ(X, Y)$ is true if $Y$ is the next integer after $X$) define relations between pairs of objects.
As customary in most ILP work, we focus on a function-free fragment\footnote{A \emph{fragment} of a logical language is a subset of this language obtained by imposing syntactical restrictions on it.} of FOL.

An \textit{atom} is a predicate applied to a tuple of arguments, either constants or variables; it is called a \textit{ground atom} if all its arguments are constants.
Well-formed FOL formulas are defined recursively as combinations of atoms with logic connectives (e.g., negation, conjunction, disjunction, implication) and existential or universal quantifiers \cg{(e.g., $\exists, \forall$)}.
\textit{Clauses} are a special class of formulas, which can be written as a disjunction of \textit{literals}, which 
are atoms or their negations.
\textit{Horn clauses} are clauses with at most one positive literal, while \textit{definite Horn clauses} contain exactly one.
In the context of ILP, definite clauses play an important role since they can be \cg{re-written as \textit{rules}}:
%cf there formation rule https://en.wikipedia.org/wiki/First-order_logic
%Claire: formation rules define how well-formed FOL formulas are generated.
\begin{equation}
    \mathsf H \leftarrow \mathsf B_1 \land \mathsf B_2 \land \dots \land \mathsf B_k
\end{equation}
where $\mathsf H$ is called the head atom and $\mathsf B_i$'s the body atoms.
The variables appearing in the head atom are instantiated with a universal quantifier, while the other variables in the body atoms are existentially quantified\footnote{This rule can be read as: if, \cg{for a certain grounding,} all the body atoms are true, then the head atom is also true.}.

\textit{Inductive Logic Programming} (ILP) aims at finding a set of rules such that all the positive examples and none of the negative ones of a target predicate are \textit{entailed} by both these rules and some background knowledge expressed in FOL.
The background knowledge may contain ground atoms, called \textit{facts}, but also some rules.
%(i.e., are logical consequences of): we should expect reader be familiar with entail, no?

\textit{Forward chaining} can be used to chain rules together to deduce a target predicate \cg{valuation} from some facts.
Consider the \textit{Even-Succ} task as an example. 
Given background facts \{$Zero(0)$, $Succ(0, 1)$, $Succ(1, 2)$\} and rules:
\begin{equation} \label{eq:evenSucc_solution}
\begin{split}
Even(X)&\leftarrow Zero(X)\\
Even(X)&\leftarrow Even(Y) \land Aux(Y, X)\\
Aux(X, Y)&\leftarrow Succ(X, Z) \land Succ(Z, Y),
\end{split}
\end{equation}
we can easily deduce the facts \{$Aux(0, 2)$, $Even(2)$\} by unifying the body atoms of the rules with the facts;
repeating this process, we can, iteratively, infer all even numbers.

The predicates appearing in the \cg{background} facts are called \textit{input predicates}.
Any other predicates, \cg{apart from the} target predicate, are called \textit{auxiliary predicates}.
A predicate can be \textit{extensional}, as defined by a set of ground atoms, or \textit{intensional}, defined by a set of clauses .%head of some rules.
%Thus, the input predicates are usually extensional and, \cg{in the context of predicate invention, we aim to define both auxiliary and target predicates intensionally.}
%as in context KB completion, they may not define target predicate intensionally, only its valuation from different rules

\subsection{Predicate Invention and Meta-Rules}

One important part of solving an ILP problem is \textit{predicate invention}, which consists in creating auxiliary predicates, \cg{intensionally defined, which would ultimately} help define the target predicate.
Without \textit{language bias}, the set of possible rules to consider grows exponentially in the number of body atoms and in the \cg{maximum arity of the predicates allowed}. 
In previous work, this bias is often enforced using \textit{meta-rules}, also called \textit{rule templates}, which \cg{restrict the hypothesis space by imposing syntactic constraints.}
\pw{The \emph{hypothesis space} is generated by the successive applications of the meta-rules \cg{on the predicate symbols from} the background knowledge.
}
%constrained the form of the rules that are generated.

A meta-rule corresponds to a second-order clause with predicate variables. For instance, the \textit{chain meta-rule} is:
\begin{equation} \label{eq:simple ex meta-rule}
 \mathrm H(X, Y) \leftarrow \mathrm B_1(X, Z) \land \mathrm B_2(Z, Y),
\end{equation}
where $\mathrm H$, $\mathrm B_1$, and $\mathrm B_2$ correspond to predicate variables.
%Variable $\mathrm H$ can be replaced by an auxiliary or target predicate, and $\mathrm B_1$ and $\mathrm B_2$ by input or auxiliary predicates.
% actually, meta-rule can be defined for any predicate, but indeed, in the case of predicate inventionm consistent meta rules are the one as above...
%(or any predicates if recursion is allowed)
% in theory, H may also be input predicate unless assume them independent? May shorten above

% As an illustration of a differentiable approach to predicate invention,
%$\partial$ILP \citep{evans_learning_2017} uses task-specific meta-rules to generate a subset of possible rules with two atoms in their body with predicates of arity up to 2.
%Solving an ILP task with this model corresponds to learning weights to select rules among this subset.
%Since this subset needs to be generated, the method does not scale well both in terms of memory and computation.
%Inspired by the architecture of Neural Logic Machines (NLM) \citep{dong_neural_2019}, 
% Said this in the related work

\paragraph{LRI Model}
\cg{\pw{We recall} the differentiable approach to predicate invention proposed by \citet{campero2018logical}}.
\cg{Their model, Logical Rule Induction (LRI), learns an embedding for each predicate and each meta-rule atom};
then, predicates and atoms of meta-rules are matched via a soft unification \pw{technique}.
For any predicate $P$, let $\bm\theta_P\in\mathbb{R}^d$ denotes its $d$-dimensional\footnote{In \citet{campero2018logical}, $d$ equals the number of predicates.} embedding. 
The embeddings attached to a meta-rule like \eqref{eq:simple ex meta-rule} with one head ($\mathrm H$) and two body ($\mathrm B_1, \mathrm B_2$) predicate variables are
$(\bm\theta_{\mathrm H}, \bm\theta_{\mathrm B_1}, \bm\theta_{\mathrm B_2})\in \mathbb{R}^d \times \mathbb{R}^d \times \mathbb{R}^d$. 
The soft unification is based on cosine to measure the degree of similarity of embeddings of predicates and meta-rule atoms. 
For example, $\alpha_{P\mathrm H} = \cos(\bm\theta_{P}, \bm\theta_{\mathrm H})\in [0, 1]$ is the unification score between predicate $P$ and head predicate variable $\mathrm H$: a higher unification score indicates a higher belief that $P$ is the correct predicate for $\mathrm H$. 
Similarly, unification scores are computed 
%\cg{for each body atom of a meta-rule (matched with each predicate).}
for any pair of predicate and meta-rule atom. 

In this model, a ground atom $\mathsf P = P(s, o)$ is represented as $(\bm\theta_P, s, o, v_{\mathsf P})$, where $\bm\theta_P$ is the embedding of predicate $P$, 
$s$ and $o$ are respectively the subject and object constants, and $
v_{\mathsf P}\in[0,1]$ is a valuation measuring the \cg{belief that $\mathsf P$ is} true. 
In ILP tasks, the valuations are initialized to 1 for background facts and are otherwise set to 0, reflecting a \textit{closed-world assumption}.

\pw{For better clarity, let use meta-rule \eqref{eq:simple ex meta-rule} as a running example to present how one inference step is performed.
}
%The following equations can easily be adapted for a different meta-rule.
For an intensional predicate $P$, the valuation of a corresponding ground atom $\mathsf P=P(x, y)$ with respect to a meta-rule $\mathfrak{R}$ of the form \eqref{eq:simple ex meta-rule} and two ground atoms, $\mathsf P_1 = P_1(x, z)$ and $\mathsf P_2 = P_2(z, y)$, is computed as follows:
\begin{equation}
    v(\mathsf P, \mathfrak{R}, \mathsf P_1, \mathsf P_2) = \left( \alpha_{P\mathrm H} \!\cdot \alpha_{P_1\mathrm B_1} \!\cdot \alpha_{P_2\mathrm B_2} \right) \cdot \left(v_{\mathsf P_1} \!\cdot v_{\mathsf P_2}  \right).
\label{equ:campero_valuation} 
\end{equation}
\cg{The \pw{term in the} first bracket\pw{s}} measures how well the \cg{predicate tuple} $(P, P_1, P_2)$ matches the meta-rule, while \cg{the \pw{term in the second} brackets} corresponds to a fuzzy AND.
\pw{Note that \eqref{equ:campero_valuation}, defined as a product of many terms in $[0, 1]$, leads to an underestimation issue.}
The valuation of atom $\mathsf P$ with respect to meta-rule $\mathfrak{R}$ is \cg{then computed as}\footnote{\cg{The max over ground atoms is taken both over possible existential variable (e.g., $Z$ above), and over predicates $P_1, P_2 \in \Predicates$.}}:
\begin{equation}
    v(\mathsf P, \mathfrak{R}) = \max_{\mathsf P_1, \mathsf P_2} v(\mathsf P, \mathfrak{R}, \mathsf P_1, \mathsf P_2).
\end{equation}
%The max corresponds to an evaluation with respect to the existential quantifier over variable $Z$.
Since $P$ can be matched with the head of several meta-rules, the new valuation of $\mathrm P$ is, after one inference step:
\begin{equation}
\label{equ:one_step_inference_campero}
    v(\mathsf P) = \max \left( v_{old}(\mathsf P), \max_{\mathfrak{R}} v(\mathsf P, \mathfrak{R}) \right),
\end{equation}
where $v_{old}(\mathsf P)$ is the previous valuation of $\mathsf P$.
\pw{The $\max$ over meta-rules allows an intensional predicate to be defined as a disjunction.}
After $n_I$ steps of forward chaining, the model uses binary cross-entropy as loss function to measure the difference between inferred values and target values.% for the target predicate

\section{Hierarchical Rule Induction}\label{sec:model}
%%%%%%%%%%%%%%%%%%%%%%%%%%%%%%%%%%%%%%%%%%%%%%%%%%

Let us introduce our model \ours and our generic meta-rules set, alongside some theoretical results about its expressivity.

\subsection{Proposed Model}\label{sec:proposed_model}

\pw{Building on LRI \citep{campero2018logical}, our model includes several innovations, 
\cg{backed both by theoretical and experimental results:}
proto-rules,
incremental prior,
hierarchical prior, and
improved model inference.}

\begin{figure}[!ht]%tbp?
   \centering
        \includegraphics[width=1.0\columnwidth]{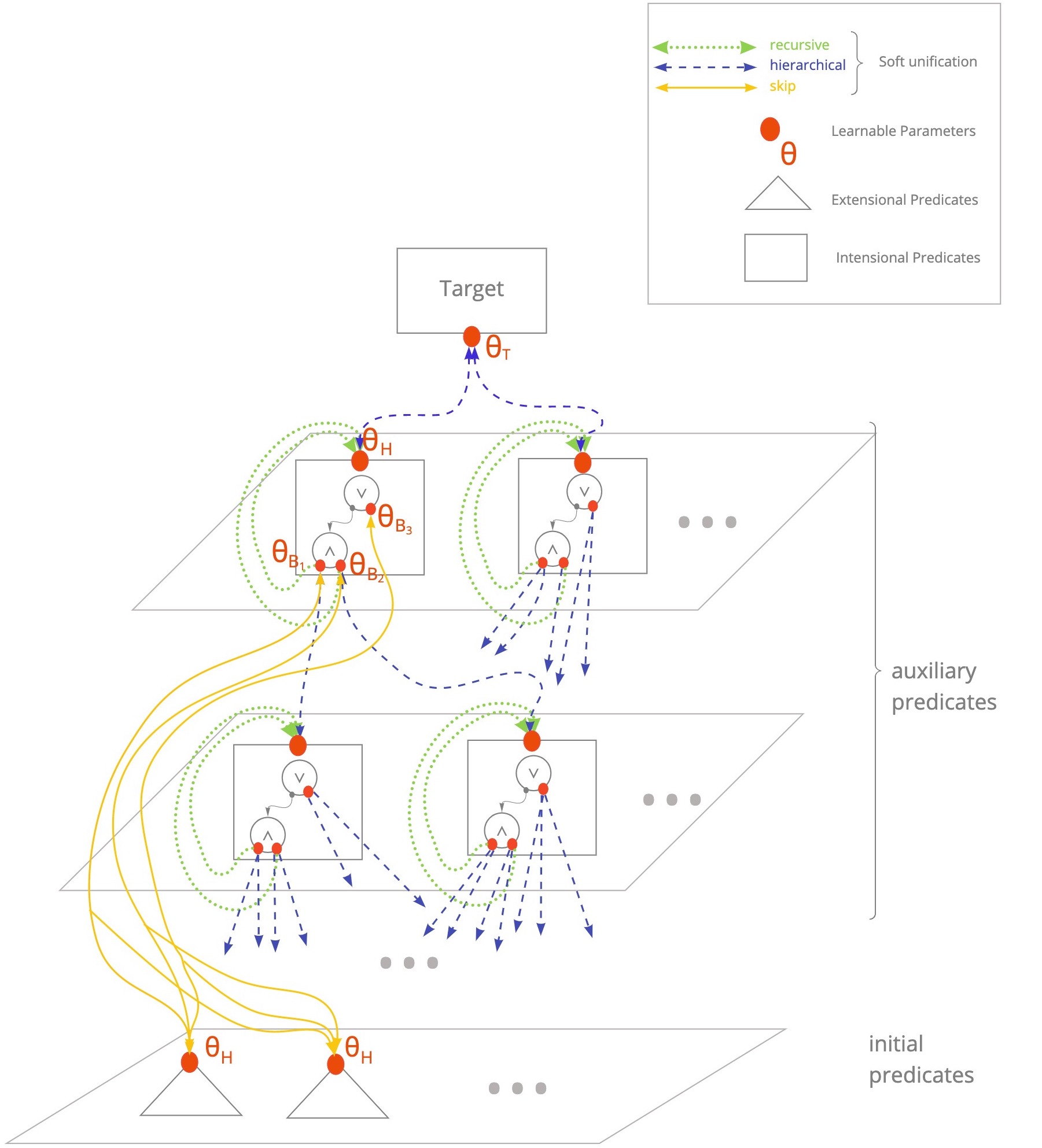} 
    \caption{Hierarchical Model}
    \label{fig:hierarchical_model}
\end{figure}

The \pw{priors} \cg{reflect} the hierarchical and progressive structure arguably present in the formation of human conceptual knowledge.
\cgnew{\ours is illustrated in Figure~\ref{fig:hierarchical_model}. The boxes, spread in different layers, denote auxiliary predicates, which embody an asymmetrical disjunction of a conjunction of two atoms with a third body predicate. The arrows represent the soft unification between one body predicate and the heads of lower-level predicates, following (\ref{eq:cos_softmax}).}

\paragraph{\pw{Proto-rules}}

We first define the notion of proto-rules, which implicitly correspond to sets of meta-rules.
A \textit{proto-rule} is a meta-rule \cg{adaptive}\footnote{
It amounts to embed some lower-arity predicates as higher-arity predicates.
}
%\cg{Formally, it corresponds to a projection onto the space of higher arity predicates.}} to multiple arity of its body atoms}
%where some variables in the body atoms are optional 
(see Appendix~\ref{appendix:meta_rules} for a formal definition).
For instance, \eqref{eq:simple ex meta-rule} \cg{extended as} a proto-rule could be written as:
\begin{equation} \label{eq:simple ex proto-rule}
\mathrm H(X, Y) \leftarrow \overline{\mathrm B}_1(X, Z) \land \overline{\mathrm B}_2(Z, Y),
\end{equation}
where the $\overline{\mathrm B}_i$'s correspond to predicate variables, \cg{of arity $2$ or $1$}, \pw{where} the second argument is optional.
% \begin{definition}
% A proto-rule is an extension of a second-order Horn clause of the form: $A_0\leftarrow \overline{P_1}(T_1,...,T_{n_{1}}) \land... \land  \overline{P_m}(T_1,...,T_{n_{m}})$,
% where $P_i$'s are predicate symbols or a second-order variable that can be substituted by a predicate symbol, and each $T_i$ is either a constant symbol or a first-order variable that can be substituted by a constant symbol.
% \end{definition}
For instance, \eqref{eq:simple ex proto-rule} can be instantiated as the following rule:
\begin{equation} \label{eq:simple ex rule}
 P(X, Y) \leftarrow P_1(X) \land P_2(Z, Y)
\end{equation}
where $P$, $P_2$ are binary predicates, and
$P_1$ is a unary predicate \cg{implicitly viewed as the binary predicate $\overline{P_1}$, where $\forall z, \overline{P_1}(x, z) \coloneqq P_1(x)$.}

We propose a minimal and expressive set of proto-rules in Section~\ref{sec:set}. However, \ours  can be instantiated with any set $\PR$ of proto-rules.
The choice of $\PR$ defines the language bias used in the model.

\paragraph{Incremental Prior}

To reinforce the incremental aspect of predicate invention, in contrast to LRI, each \pw{auxiliary} predicate is directly associated with a unique \pw{proto-}rule defining it.
\pw{This has two advantages.
First, it partially addresses the underestimation issue of \eqref{equ:campero_valuation} since it amounts to setting $\alpha_{P\mathrm H} = 1$.
Second, it reduces computational costs since the soft unification for a rule can be computed only by matching the body atoms\footnote{The $\max$ over \pw{meta-}rules in  \eqref{equ:one_step_inference_campero} is not needed anymore.% since an intensional predicate is identified to a \pw{proto-}rule
}.
\cg{In order} \pw{to preserve expressivity}\footnote{Since identifying intensional predicates to proto-rules prevent them to have disjunctive definitions.},
we can} re-introduce disjunctions in the model
\pw{by considering proto-rules with disjunctions in their bodies (see Section~\ref{sec:set});
\cg{e.g.}, \eqref{eq:simple ex proto-rule} \pw{could be extended to}:
\begin{equation} \label{eq:simple ex proto-rule extended}
\mathrm H(X, Y) \leftarrow \big(\overline{\mathrm B}_1(X, Z) \land \overline{\mathrm B}_2(Z, Y)\big) \lor \overline{\mathrm B}_3(X, Y).
\end{equation}
}

\paragraph{Hierarchical Prior}

A hierarchical architecture is enforced by organizing auxiliary predicates in successive layers, from layer $1$ to the max layer $n_L$,\cgnew{ as illustrated   Figure~\ref{fig:hierarchical_model}} . %, referred to as their depths. 
% d already refers to the dimension of the embedding space
\pw{Layer $0$ contains all input predicates.}
\pw{For each layer $l=1, \ldots, n_L$}, each \pw{proto}-rule\footnote{\pw{Alternatively, several auxiliary predicates could be generated per proto-rules. For simplicity, we only generate one and control the model expressivity with only one hyperparameter $n_L$.}} \cg{in} $\PR$ generates one auxiliary predicate. % while the \pw{last layer $n_L$ contains only \cg{one} auxiliary predicate \cg{per} \cg{proto}-rule in $\PR$ of the same arity as the target predicate}.

\pw{Lastly, the target predicate is \cg{matched with the} auxiliary predicates in layer $n_L$ only.}

With this hierarchical architecture, an auxiliary predicate $P$ at layer $\ell$ can be only defined from a set $\Predicates_{\ell}^\downarrow$ of predicates at a layer lower or equal to $\ell$.
This set can be defined in several ways depending on whether recursivity is allowed.
We consider three cases: no recursivity, iso-recursivity, and full recursivity.
For the no recursivity (resp. full recursivity) case, $\Predicates_{\ell}^\downarrow$ is defined as the set $\Predicates_{<\ell}$ (resp. $\Predicates_{\le\ell}$) of predicates at layer strictly lower than (resp. lower or equal to) $\ell$.
For iso-recursivity, $\Predicates_{\ell}^\downarrow$ is defined as $\Predicates_{<\ell} \cup \{P\}$.
%We detail different recursive designs in the Appendix\TODO{~\ref{appendix:meta_rules}}.

Enforcing this hierarchical prior has two benefits.
First, it imposes a stronger language bias, which facilitates learning.
Second, it also reduces computational costs since the soft unification does not need to consider all predicates.
%\pw{If recursive definitions are not allowed, intensional predicates at \pw{layer $\ell$} can only be defined from strictly lower-\pw{layer} predicates. 
%therwise, they can also be defined from predicates at the same \pw{layer}.}
%The max depth $\textbf{D}$ is then a hyper-parameter which shall be no less than the depth of the simplest solution to a task.

\paragraph{Improved Model Inference}

\pw{We improve LRI's inference technique with a soft unification computation that reduces the underestimation issue in \eqref{equ:campero_valuation}, which is due to the product of many values in $[0, 1]$.
\cg{For clarity\pw{'s sake}, we illustrate the inference} with proto-rule~\eqref{eq:simple ex proto-rule extended};
although the equations can be straightforwardly adapted to any proto-rule.
One inference step} in our model is formulated as follows\footnote{\pw{It generalizes \eqref{equ:campero_valuation}-\eqref{equ:one_step_inference_campero}, under the incremental prior.}}:
\begin{equation}\label{equ:one_step_inference}
\arraycolsep=1pt\def\arraystretch{1.5}
\begin{array}{ll}
   v_{and} & = \textsc{pool}_{\mathsf P_1, \mathsf P_2} \left( \alpha_{P_{1}\mathrm B_1} \cdot \alpha_{P_{2}\mathrm B_2} \cdot \textsc{and}[ v_{\mathsf P_1}, v_{\mathsf P_2}] \right) \\
    v_{or}   & = \textsc{or} \left[ v_{and} , \textsc{pool}_{\mathsf P_3} \left( \alpha_{P_{3}\mathrm B_3} \cdot v_{\mathsf P_3}\right) \right] \\
    v &=\textsc{merge} \left( v_{old}, v_{or}\right),
\end{array}
\end{equation}
where $v$ (resp. $v_{old}$) denotes the new (resp. old) valuation \pw{of a grounded auxiliary predicate $\mathsf P$}. 
\pw{For an auxiliary predicate at layer $\ell$, the $\textsc{pool}$ operation is performed over both predicates $P_1, P_2, P_3 \in \Predicates_{\ell}^\downarrow$ and the groundings that are compatible to $\mathsf P$.
\cgnew{However, before pooling, in presence of an existential quantifier as in proto-rule (\ref{eq:simple ex proto-rule extended}), a max is applied to the corresponding dimension of the tensor.
}
%(i.e., the same variable is replaced by the same constant)
The valuation $v^{t}$ of the target predicate (with variable denoted $\mathrm P^t$) is computed as a $\textsc{merge}$ of its previous valuation $v^{t}_{old}$ and a \cg{$\textsc{pool}$ over atoms \pw{at layer $n_L$}:}}

\begin{equation}\label{equ:one_step_inference_tgt}
v^{t} =\textsc{merge} \left( v^{t}_{old}, \textsc{pool}_{\mathsf P_1} \left( \alpha_{P_{1}\mathrm P^t}  \cdot v_{\mathsf P_1} \right) \right).
\end{equation}
%\pw{where $\mathrm P^t$ corresponds to a predicate variable for the target predicate.}

\pw{The underestimation issue is alleviated in two ways.
First, the $\textsc{pool}$ operation is implemented as a sum, $\textsc{and}$ as $\min$, $\textsc{or}$ as $\max$, and $\textsc{merge}$ as $\max$.
Second, the unification scores $\alpha_{P_i\mathrm B_i}$'s based on cosine are renormalized with a softmax transformation: 
\xf{
\begin{equation} \label{eq:cos_softmax}
    \alpha_{P_i\mathrm B_i} = \frac
    {\exp( \cos(\bm\theta_{P_i}, \bm\theta_{\mathrm B_i})/\tau)}
    {\sum_{P_j} \exp(\cos(\bm\theta_{P_j}, \bm\theta_{\mathrm B_i})/\tau)}, 
\end{equation}}}%
where $P_j \in \Predicates_{\ell}^\downarrow$ and hyperparameter $\tau$, called temperature, controls the renormalization.
\pw{Score $\alpha_{P_{1}\mathrm P^t}$ can be computed in a similar way with embedding $\bm \theta_{\mathrm P^t}$.}
%While diverse choices and combinations of fuzzy $\textsc{and}$, fuzzy $\textsc{or}$, similarity measure $\alpha$, merging and pooling functions have been explored, the experimental results led us to adopt max as merging, sum as pooling, min resp. max as fuzzy and resp. or.\\
%Regarding the unification scores $\alpha_{P_{i}B_{i}}$, we compose the cosine similarity with a \textit{Gumbel-Softmax}. The softmax is normalising the score along the first predicate dimension, while the role of the Gumbel noise is discussed in the next section.
These choices have been further experimentally validated (see Appendix~\ref{appendix:operation choices}).

\pw{Equation~\eqref{equ:one_step_inference} implies that the computational complexity of one inference step at layer $\ell$ if there is no recursion is $\mathcal O(n_L \times |\Predicates_{\ell}| \times |\Predicates_{\ell}^\downarrow|^2 \times |\Constants|^2)$ where
$\Predicates_{\ell}$ is the set of predicates at layer $\ell$ and $\Predicates_{\ell}^\downarrow$ is the set of predicates available for defining an auxiliary predicate at layer $\ell$.
Its spatial complexity at layer $\ell$ is $\mathcal O(n_L \times |\Predicates_{\ell}| \times |\Constants|^2)$.
}

\pw{The inference step described above is iterated $n_I$ times, % (hyperparameter), 
updating the valuations of auxiliary and target predicates.
Note that when recursive definitions of predicates are allowed, it could be conceivable to iterate this inference step until convergence.
A nice property of this inference procedure is that it is parallelizable, which our implementation on GPU takes advantage of.}

\subsection{Generic Set of Proto-Rules}\label{sec:set}

\begin{figure}[tbp]
    \centering
    \begin{tabular}{@{}ccc@{}}
        \includegraphics[width=0.15\columnwidth]{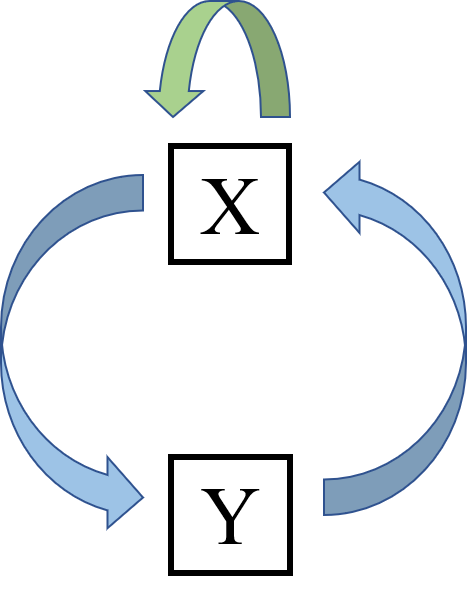} &  \includegraphics[width=0.15\columnwidth]{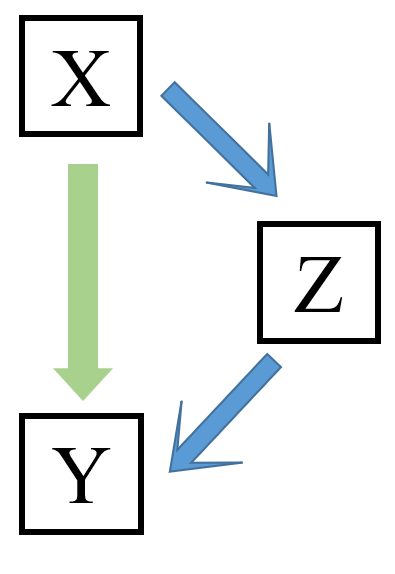} &  \includegraphics[width=0.15\columnwidth]{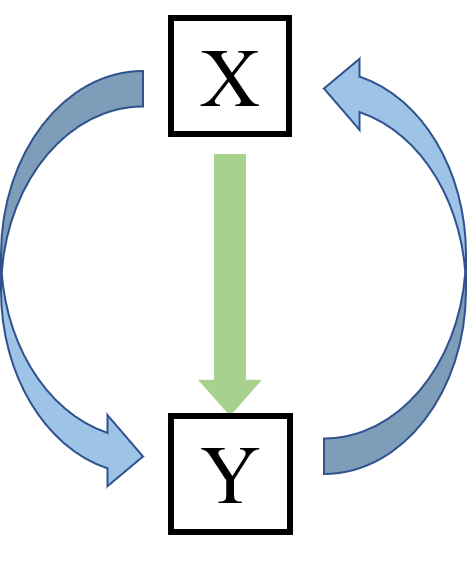} \\
        
        (a) Meta-rule $\mathfrak{A}$ & (b) Meta-rule $\mathfrak{B}$ & (c) Meta-rule $\mathfrak{C}$
    \end{tabular}
    
    \caption{Three types of meta-rules: \pw{boxes represent variables, green arrows represent relations for head atoms and blue arrows represent relations for body atoms}}
    \label{fig:rule_templates}
\end{figure}

Although \ours is generic, we design a small set  of expressive proto-rules to instantiate our model.
We introduce first the following set of proto-rules $\PR_0$ (see Figure~\ref{fig:rule_templates}):
\begin{equation*} \label{eq:template_102}
\arraycolsep=1pt\def\arraystretch{1.5}
%\PR*:=
\left\lbrace 
\begin{array}{llllll}%cccccccc}
\mathfrak{A}: & \mathrm H (X) &\leftarrow  &  \overline{\mathrm B}_1(X,Y) & \land & \overline{\mathrm B}_2(Y,X)\\
\mathfrak{B}: & \mathrm H(X, Y)& \leftarrow &  \overline{\mathrm B}_1(X, Z) & \land& \overline{\mathrm B}_2(Z, Y)\\
\mathfrak{C}:& \mathrm H (X,Y) & \leftarrow  & \overline{\mathrm B}_1(X,Y)  & \land & \overline{\mathrm B}_2(Y,X)
%Inv: &  H(X, Y) & \leftarrow &  \overline{F}(Y, X) & & & & 
\end{array}\right\rbrace%\quad ,
\end{equation*}
where the $\overline{\mathrm B}_i$'s correspond to predicate variables where the second argument is optional \cg{(see Section~\ref{sec:proposed_model})}.

To support our design, we \cg{analy\pw{z}ed} the expressivity of $\PR_0$, \cg{by investigating} which fragment of first-order logic can be generated by $\PR_0$ from a set of predicates $\Predicates$.
\cg{Because of space constraints, we only present here our main result}, 
\pw{where we assume that $\mathcal P$ contains the zero-ary predicate $\mathtt{True}$ and the equality symbol}:
\begin{theorem}\label{thm:main_theorem}
%\begin{enumerate}
%    \item[$(i)$] Assuming $ \mathtt{True} \in \Predicates_0$, the hypothesis space generated by $\PR_0$ from $\Predicates$ encompasses the set of duplicate-free function-free definite Horn clause fragment $\mathcal F_{\Predicates, \leq 2 }^{\lbrace 1,2 \rbrace}$ composed of \pw{clauses} with \pw{at most} 2 body atoms involving unary and binary predicates in $\Predicates$.
%\item[$(ii)$] Assuming $ \mathtt{True}, \mathtt{Equal} \in \Predicates_0$, t
    The hypothesis space generated by $\PR_0$ from $\Predicates$ \pw{is} exactly the set of function-free definite Horn clause fragment $\mathcal F_{\Predicates, \leq 2}^{\lbrace 1,2 \rbrace}$ composed of clauses with \pw{at most} two body atoms involving unary and binary predicates in $\Predicates$.
%\end{enumerate}
\end{theorem}
We refer the reader to Appendix~\ref{appendix:meta_rules} for all formal definitions, proofs and further theoretical results.
\pw{This results implies that our model with $\PR_0$ can potentially solve perfectly any ILP task whose target predicate is expressible in $\mathcal F_{\Predicates, \leq 2}^{\lbrace 1,2 \rbrace}$ with a large enough max layer $n_L$.}

However, to potentially reduce the max layer $n_L$ and facilitate the definition of recursive predicates, we extend $\PR_0$ to the set \pw{$\PR_0^\lor$} by including a disjunction with a third atom in all the proto-rules: 
\begin{equation*}  \hspace*{-0.3cm} \label{eq:template_102}
\arraycolsep=1pt\def\arraystretch{1.5}
%\PR*:=
\left\lbrace 
\begin{array}{llllllll}%cccccccc}
\mathfrak{A}_*: & \mathrm H (X) &\leftarrow  & \left( \overline{\mathrm  B}_1(X,Y) \right. & \land & \left. \overline{\mathrm B}_2(Y,X) \right) & \vee & \overline{\mathrm B}_3(X,T)\\
\mathfrak{B}_*: & \mathrm H (X, Y)& \leftarrow & \left( \overline{\mathrm B}_1(X, Z)\right.  & \land&  \left.\overline{\mathrm B}_2(Z, Y)\right) & \vee & \overline{\mathrm B}_3(X, Y)\\
\mathfrak{C}_*:& \mathrm H (X,Y) & \leftarrow  & \left( \overline{\mathrm B}_1(X,Y) \right. & \land &\left. \overline{\mathrm B}_2(Y,X)\right) & \vee &  \overline{\mathrm B}_3(X,Y)\\
%Inv: &  H(X, Y) & \leftarrow &  \overline{F}(Y, X) & & & & 
\end{array}\right\rbrace%\quad ,
\end{equation*}
\pw{Moreover, as} we have observed a certain redundancy to be beneficial to the learning, we incorporate the permutation rule $\mathfrak{I}$ in our experiments:
\begin{equation*}
\PR_*=\PR_0^\lor \cup  \lbrace \mathfrak{I}:  H(X, Y)  \leftarrow \overline{F}(Y, X) \rbrace
\end{equation*}
This small set of protorules \pw{$\PR_*$}, which has the same expressivity as $\PR_0$ \pw{characterized \cg{in Theorem \ref{thm:main_theorem}}, is used to instantiate our model in all our experiments.}

%%%%%%%%%%%%%%%%%%%%%%%%%%%%%%%%%%%%%%%%%%%%%%%%%%
\section{Proposed Training Method}\label{sec:training}
%%%%%%%%%%%%%%%%%%%%%%%%%%%%%%%%%%%%%%%%%%%%%%%%%%

\paragraph{Training}

We train our model with a fixed number $n_T$ of \pw{training} iterations. 
For each iteration, the model is trained using one training instance.
During each iteration, we add \pw{a geometrically-decaying Gaussian-distributed} noise to the embeddings for both predicates and rules to avoid local optima\pw{, similarly to LRI \citep{campero2018logical}}. 
Then the improved inference procedure described in \eqref{equ:one_step_inference} is performed with these noisy embeddings for a number \pw{$n_I$} of steps to get the valuations for all predicates.
%The number of inference steps is a hyper-parameter named train-steps in our implementation.
%
After \pw{these $n_I$} inference steps, the binary cross entropy (BCE) loss of the ground-truth target predicate valuation and the learned target predicate valuation is calculated:

\pw{%
\begin{equation}\label{eq:bceloss}
    \sum_{x,y} \shortminus G^t_{x, y}\log\big(v(\mathsf P^t_{x,y})\big) - (1 \shortminus G^t_{x, y})\log\big(1\shortminus v(\mathsf P^t_{x, y})\big),
\end{equation}%
where the sum is over all pairs of objects $x,y$ in one training instance, $G^t_{x, y}\in\{0,1\}$ is the ground truth value of atom $\mathsf P^t_{x, y} = P^t(x,y)$.}
To encourage \pw{the} unification scores to be closer to $0$ or $1$, an extra regularisation term %$\sum \mbox{unifs}\times(1-\mbox{unifs})$ 
is added to the BCE loss to update the embeddings:
\pw{%
\begin{equation}
 +\lambda\sum_{P, \mathrm Q} \alpha_{P\mathrm Q}\big(1-\alpha_{P\mathrm Q}\big).
\end{equation}
where hyperparameter $\lambda$ controls the regularization weight and the sum is over all predicates $P$ that can be match\cg{ed} with a predicate variable $\mathrm Q$ appearing in some meta-rules.}

\begin{table*}[t]
\centering
\begin{tabular}{cccccccc}
\toprule
\multirow{2}{*}{Task} & \multirow{2}{*}{$|I|$} & \multirow{2}{*}{Recursive} & \multirow{2}{*}{$\partial$ILP} & \multirow{2}{*}{LRI} & \multicolumn{3}{c}{Ours}                        \\
\cmidrule{6-8}
                      &                      &                            &                           &                           & train & soft evaluation & symbolic evaluation \\
                      \midrule
Predecessor           & 1                    & No                         & 100                       & 100                       & 100   & 100               & 100                 \\
Undirected Edge       & 1                    & No                         & 100                       & 100                       & 100   & 100               & 100                 \\
Less than             & 1                    & Yes                        & 100                       & 100                       & 100   & 100               & 100                 \\
Member                & 1                    & Yes                        & 100                       & 100                       & 100   & 100               & 100                 \\
Connectedness         & 1                    & Yes                        & 100                       & 100                       & 100   & 100               & 100                 \\
Son                   & 2                    & No                         & 100                       & 100                       & 100   & 100               & 100                 \\
Grandparent           & 2                    & No                         & 96.5                      & 100                       & 100   & 100               & 100                 \\
Adjacent to Red       & 2                    & No                         & 50.5                      & 100                       & 100   & 100               & 100                 \\
Two Children          & 2                    & No                         & 95                        & 0                         & 100   & 100               & 100                 \\
Relatedness           & 2                    & Yes                        & 100                       & 100                       & 100   & 100               & 100                 \\
Cyclic                & 2                    & Yes                        & 100                       & 100                       & 100   & 100               & 100                 \\
Graph Coloring        & 2                    & Yes                        & 94.5                      & 0                         & 100   & 100               & 100                 \\
%Length                & 2                    & Yes                        & 92.5                      & 100                       & 20    & 0                 & 0                   \\
%Even-Odd              & 2                    & Yes                        & 100                       & 100                       & 40    & 40                & 40                   \\
Even-Succ2            & 2                    & Yes                        & 48.5                      & 100                       & 40    & 40                & 40                  \\
Buzz                  & 2                    & Yes                        & 35                        & 70                        & 100   & 40                & 40                  \\
Fizz                  & 3                    & Yes                        & 10                        & 10                        & 0     & 0                 & 0  \\
\bottomrule
\end{tabular}
\caption{Percentage of successful runs among 10 runs. \xf{$|I|$} is the smallest number of intensional predicates needed. Recursive means whether or not the solution need\pw{s} to learn recursive rules.}
\label{table:acc-result}
\end{table*}

In addition, to further help with convergence to an interpretable solution and avoid local optima, we replace during training the softmax transformation by a variant of Gumbel-softmax \citep{jang2017categorical,Maddison_Concrete_2017}.
Commonly, it amounts to injecting a Gumbel noise $G_j$ in the softmax, i.e., 
in \eqref{eq:cos_softmax}, we replace each $\cos(\bm\theta_{P_j}, \bm\theta_{B_i})$ by $\cos(\bm\theta_{P_j}, \bm\theta_{B_i}) + G_j$ where 
Gumbel noise $G_j = -g \log \left( -\log \left(U_j \right)  \right)$ is obtained by sampling $U_j$ according to a uniform distribution. 
\cgnew{However, we rely on a variant of this noise: $\tilde{G}_j = -g \log \left( -\log \left(U_j \right)  \right)$, which is related as $\tilde{G}_j=\frac{ \log (- \log g) }{g}G_{j}$. This choice has been experimentally favored. We discuss it further in the Appendix. \TODO{Gumbel Noise}}
Gumbel scale $g \in (0, +\infty)$ is linearly decreased during training.
%In our \pw{experiments}, we use a fixed temperature and a linearly-decaying Gumbel noise. 
\cg{In contrast to the low-level parameter-noise applied directly to the embeddings, the Gumbel noise may be understood as a higher-level noise, enacted on the similarity coefficients themselves.}
% as the training iterations increase.

% We already use $d$ for the embedding dimension.
%The softmax normalizes the score along the first predicate dimension, while the role of the Gumbel noise is discussed in the next section.

\paragraph{Convergence to an Interpretable Model}

The passage from our soft model to a symbolic model may be realized by taking the limit of the softmax temperature in the unification score to zero, %$T\rightarrow 0$, where $T$ is the temperature of the softmax in the unification score, 
or equivalently, switching to an argmax in the unification score; i.e., for each \pw{proto-rule}, the final learned rules can be interpreted by assigning head and body atoms to the predicates that obtain the highest unification score. 

For instance, at a \pw{layer $\ell$}, the \pw{extracted} symbolic rule \eqref{eq:simple ex rule} could be extracted from \pw{proto-rule \eqref{eq:simple ex proto-rule}} if 
$P$ is the auxiliary predicate associated to that proto-rule at layer $\ell$ and 
\begin{equation} \label{equation:interpret_rules}
\left\{
\begin{array}{ccc}
  P_{1}   & =& \arg\max_{P\in \mathcal{P}^\downarrow_{\ell}}\: %cos(\theta_{b_1}, \theta_q), \\
  \alpha_{P\mathrm B_1} \\
  P_{2}   & =& \arg\max_{P\in \mathcal{P}^\downarrow_{\ell}}\: %cos(\theta_{b_2}, \theta_q). \\
  \alpha_{P\mathrm B_2}
\end{array}
\right.
\end{equation}
%As seen through this example, our templates are not restricting the arity of the body predicates.
%Note the presence of $\mathcal{P}^{\leq d}$ denoting the subset of predicates of \pw{layer} smaller than \pw{$\ell$}, which reflects the hierarchical prior of our model.

Then we can interpret the solution by successively \cg{unfolding the logical formula extracted for} each involved predicate, starting from the target predicate.

%%%%%%%%%%%%%%%%%%%%%%%%%%%%%%%%%%%%%%%%%%%%%%%%%%
\section{Experimental Results}\label{sec:experiments}
%%%%%%%%%%%%%%%%%%%%%%%%%%%%%%%%%%%%%%%%%%%%%%%%%%

We evaluated our model on multiple domains and compared with relevant baselines.
In the main paper, for space constraints, we only present two sets of experiments.
First, on standard ILP tasks, we show that our model is competitive against other neuro-symbolic methods in terms of performance, but also computational/space costs. 
Second, on a large domain from Visual Genome \citep{krishna_visual_2017}, we show that our model can scale to a large multi-task problem and is superior to NLIL \citep{yang_learn_2019}. 

Our results are averaged over several runs with different random seeds.
In each run, the model is trained over a fixed number of iterations. 
Hyper-parameter details are given in Appendix~\ref{appendix:Hyper-param}, 
and additional experiments (e.g., choice of operators, limitations of LRI, reinforcement learning) are discussed in Appendices~\ref{appendix:ILP_further_results} and \ref{appendix:other}.

For simplicity and consistency, we instantiate our model with $\PR_*$, with full recursivity in all the tasks except \zj{\xf{V}isual \xf{G}enome (no recursivity)}.

Note the performances can be improved and/or learning can be accelerated if we customize $\PR_*$ to a task \cg{or restrict the recursivity}.
However, we \cg{intended} to emphasize the genericity of our approach. 
%, we did not try to further optimize the performances.

%For each run, we feed our model with randomly generated data for stochastic tasks and fixed data for deterministic tasks, similarly to \citet{dong_neural_2019}.

\subsection{ILP Tasks}

\begin{table*}[t]
\centering
\parbox{.6\linewidth}{
\centering
\begin{tabular}{@{}cccccccc@{}}
\toprule
\multirow{2}{*}{Task} & \multirow{2}{*}{\#Training} &
\multicolumn{3}{c}{\% successful runs} &           
\multicolumn{3}{c}{Training time (secs)}                        \\
\cmidrule{3-8}
                      & constants                             & NLM & DLM & Ours & NLM & DLM & Ours \\
\midrule
\multirow{2}{*}{Adjacent to Red}    & 7 & 100 & 90 & 100 & 163 & 920 & 62 \\ %213  \\
                                    & 10    & 90 & 90 & 100 & 334  & 6629  & 71 \\ %1036 \\%1624   \\

\midrule
\multirow{2}{*}{Grandparent}        & 9 & 90 & 100 & 100 & 402 & 2047 & 79 \\ %461 \\
                                    & 20    & 100 & 100 & 100 & 1441  & 3331  & 89 \\ %11104  \\
\bottomrule
\end{tabular}
\caption{Comparisons with NLM/DLM in terms of percentage of successful runs and average training times over 10 runs.}
\label{table:time-result-compare}
}
\hfill
\parbox{.32\linewidth}{
\centering
\begin{tabular}{@{}lll@{}}
\toprule
\multirow{2}{*}{Model} & \multicolumn{2}{l}{Visual Genome} \\
\cmidrule{2-3}
                       & R@1             & R@5             \\
\midrule
MLP+RCNN               & 0.53            & 0.81            \\
Freq                   & 0.40            & 0.44            \\
NLIL                   & 0.51            & 0.52            \\
\midrule
Ours                   & 0.53            & 0.60           \\
\bottomrule
\end{tabular}
\caption{R@1 and R@5 for 150 objects classification on
VG.}
\label{table:res_MT_GQA}
}
\end{table*}

We evaluated our model on the ILP tasks from \citep{evans_learning_2017}, which are detailed in Appendix~\ref{appendix:ILP tasks description}, where we also provide the interpretable solutions found by our model.
%The performances of our model compared to $\partial$ILP \citep{evans_learning_2017} and LRI \citep{campero2018logical} \pw{on a selection of these ILP tasks} are listed in Table~\ref{table:acc-result} (see Appendix~\ref{appendix:ILP_further_results} for the whole table). 

%The results of $\partial$ILP and LRI are those reported in their papers.
We compare our model with $\partial$ILP \citep{evans_learning_2017} and LRI \citep{campero2018logical} using their reported results. 
In contrast to LRI or $\partial$ILP, which use a specific set of meta-rules tailored for each task, we use the same generic and more expressive template set $\PR_*$ to tackle all tasks;
\cg{naturally, this generic approach impedes the learning for a few tasks.}
%Using a specifically-designed meta-rules is not practical since it requires a good knowledge of the form of the solution, which is generally not available.
%their performances could be improved if we use a customized or narrower set of proto-rules. % Already said above

%We evaluate the trained model with 10 new \pw{larger} datasets \zj{for stochastic tasks and with 1 new larger dataset for deterministic tasks}. 

A selection of the results on these ILP tasks are presented in Table~\ref{table:acc-result} (see Appendix~\ref{appendix:ILP_further_results}).
A run is \pw{counted} as \emph{successful} if the mean square error between inferred and given target valuations is less than \pw{1e-4} for new evaluation datasets. 
\pw{Column \emph{train} corresponds to the performance measured during training, while Column \emph{soft evaluation} resp. \emph{symbolic evaluation} to evaluation performance (no noise), using \eqref{equ:one_step_inference_tgt} \cg{resp.} the interpretable solution (see end of Section~\ref{sec:training}).}
% For instance, Length requires two proto-rules B at the same layer and full-recursivity.

We also ran our model on two other ILP tasks used by NLM \citep{dong_neural_2019} and DLM \citep{zimmer2021differentiable} to compare their performances and training times for a varying number of training constants.
Those two ILP tasks are similar to those used in \citep{evans_learning_2017}, but with different background predicates.
The results reported in Table~\ref{table:time-result-compare} show that our model is one to two orders of magnitude faster  and that it scales much better in terms of training constants, while achieving at least as good performances.

\begin{table*}[t]
\centering
\small
\begin{tabular}{ccccccc}
\toprule
\multirow{2}{*}{Initialization} & \multicolumn{2}{c}{Accuracy} & \multicolumn{2}{c}{Precision} & \multicolumn{2}{c}{R@1}                        \\
\cmidrule{2-7}
                      & soft evaluation & symbolic evaluation    & soft evaluation & symbolic evaluation   & soft evaluation & symbolic evaluation \\
                      \midrule
Random          & 0.63  & 0.49  & 0.57  & 0.5   & 0.23   & 0.38  \\
NLIL            & 0.75  & 0.6   & 0.87  & 0.75  & 0.46   & 0.58  \\
GPT2         & 0.65  & 0.45  & 0.72  & 0.66  & 0.27   & 0.5   \\
\bottomrule
\end{tabular}
\caption{Performance of different embedding initializations for single Visual Genome task.}
\label{table:single-vg-task-compare}
\end{table*}

\subsection{Visual Genome}

\cg{Our model can be applied beyond classical ILP domains, and even benefit from richer environments and semantic structure. 
Futhermore, initial predicates embeddings can be boostrapped by visual or semantic priors.
We illustrate it by applying it to the larger dataset of Visual Genome \citep{krishna_visual_2017}, or more precisely a pre-processed (less noisy) version known as the GQA \citep{Hudson2019GQAAN}. 
Similarly to \citet{yang_learn_2019}, we filtered out the predicates with less than 1500 occurrences, which lead to a KB with 213 predicates;
then, we solved ILP-like task to learn the explanations for the top 150 objects in the dataset. 
More details on task, results, baseline and metrics used are provided in Appendix~\ref{appendix:visual_genome_experiments}. 
}

We ran a first set of experiments to validate that \cg{embeddings priors may be leveraged} and compared three initialization methods: \xf{random initialization, pretrained embeddings from NLIL \citep{yang_learn_2019}, and pretrained embeddings from NLP model GPT2 \citep{Radford2019LanguageMA}.
We trained our model on a single ILP task, which consists in predicting predicate \textit{Car}. 
For this task, we further filtered the GQA dataset to keep 185 instances containing cars. 
%There were 185 data instances after the filtering and we trained on this dataset with the above three ways of embedding initialization.
%To perform a better evaluation, we computed accuracy, precision and recall for this task.
%The results are displayed in Table~\ref{table:single-vg-task-compare} and each of them was averaged over 10 runs.
Table~\ref{table:single-vg-task-compare} presents the accuracy, precision, and recall, obtained over 10 runs.
}%
\xf{
%From Table~\ref{table:single-vg-task-compare}, we see that both p
Pretrained embeddings outperform the randomly initialized ones in all metrics.
%Therefore, our model benefits from learned embeddings of predicates when trying to learn rules for new tasks, compared with random initialized embeddings.
Embeddings from NLIL, trained specifically on this dataset, expectedly yield the best results.
%Inspired by this, we further utilize the embeddings from GPT2 to a solve multitask problem with a much larger dataset.
Although, to be fair, for the next experiments, we rely on the NLP embeddings.
}

\zj{In multitask setting, we trained 150 target models together with shared background predicate embeddings, where each model corresponds to one object classification. 
%Here the background embeddings are initialized from NLP embeddings.%we say it above that we keep NLP embeddings from nnow on (need space :))
%Unlike single task, we did not filter GQA further in this part. 
The training consists of $N_{r}$ rounds.
In each round, we trained each model sequentially, by sampling for each target $n_{p}$ instances containing positive examples, and $n_{r}$ with or without positive examples.

We compared our model with NLIL \citep{yang_learn_2019} and two other supervised baselines (classifiers) mentioned in their paper: Freq, which predicts object class by checking the related relation that contains the target with the highest frequency; and MLP-RCNN, a MLP classifier trained with RCNN features extracted from object images in Visual Genome dataset.
The latter is a strong baseline because it uses visual information for its predictions while the other methods only use relational information. 
Like NLIL, we use R@1 and R@5 to evaluate the trained models. Table \ref{table:res_MT_GQA} summarizes the results. 
We see that both our method and MLP+RCNN achieve best performance for R@1. 
For R@5, we outperform Freq and NLIL. Table \ref{table:interprete_MT_rules} shows some extracted learned rules for the multitask model.}
% Freq predict object class by checking the related relation that contains the target with the highest frequency.

%\zj{
%We also checked if the fine-tuned background predicate embeddings keep similar attribution with original embeddings from GPT2. 
%}
% Examples of close predicates in  VGA: 
% on vs. over and 
% hanging from vs. attached to

\begin{table}[]
\fontsize{8}{6}\selectfont
\hspace*{-0.4cm}
\begin{tabular}{@{}cc@{}}
\toprule
Target & Rules \\
\midrule
wrist  & 
$\begin{aligned}[t] 
wrist(X) & \leftarrow \left( watch(Y) \land on(Y,X) \right) \vee False
\end{aligned}$\\
\midrule
person &
$\begin{aligned}[t] 
person(X) & \leftarrow \left( on(X, Y) \land bench(Y) \right) \vee wearing(X, T)
\end{aligned}$\\
\midrule
vase   & 
$\begin{aligned}[t] 
vase(X) & \leftarrow \left( aux(X, Y) \land flowers(Y) \right) \vee False \\
aux(X, Y) & \leftarrow \left( True \land on(Y,X) \right) \vee with(X,Y)
\end{aligned}$ \\
\midrule
logo   &
$\begin{aligned}[t] 
logo(X) & \leftarrow \left( aux_0(X) \land True \right) \vee aux_1(X) \\
aux_1(X) & \leftarrow \left( on(X, Y) \land laptop(Y) \right) \vee False\\
aux_0(X) & \leftarrow \left( on(X, Y) \land kite(Y) \right) \vee False
\end{aligned}$\\
\bottomrule
\end{tabular}
\caption{Examples of extracted rules from multi-task model}
\label{table:interprete_MT_rules}
\end{table}

%%%%%%%%%%%%%%%%%%%%%%%%%%%%%%%%%%%%%%%%%%%%%%%%%%
\section{Ethical Considerations}\label{sec:ethical}
%%%%%%%%%%%%%%%%%%%%%%%%%%%%%%%%%%%%%%%%%%%%%%%%%%

\cgnew{Regarding our last experiment choice, let us point out that the use of data coming from annotated images, as well as the use of pretrained embeddings produced by the NLP model GPT2 should be regarded with circumspection. As it as been well documented \citep{Papakyriakopoulos, Tommasi}, such data holds strong biases, which would raise ethical issues when deployed in the real-world. Our motive was merely to illustrate the performance of our model in more noisy domains typical of real-world scenarios (such as autonomous driving scenarios). We strongly advise against the use of such unbiased data for any real-world deployement.
}

%%%%%%%%%%%%%%%%%%%%%%%%%%%%%%%%%%%%%%%%%%%%%%%%%%
\section{Conclusion}\label{sec:conclusion}
%%%%%%%%%%%%%%%%%%%%%%%%%%%%%%%%%%%%%%%%%%%%%%%%%%

\cg{We presented a new neuro-symbolic interpretable model performing hierarchical rule induction through soft unification with learned embeddings;
it is initialised by a \emph{theoretically supported} small-yet-expressive set of proto-rules, which is sufficient to tackle \cgnew{many classical ILP benchmark tasks},
as it encompasses a consequent function-free definite Horn clause fragment.
% ----EXPERIMENTAL RESULT SUMMARY
Our model has demonstrated its efficiency and performance in both ILP, \pw{reinforcement learning (RL)} and richer domains against state-of-the-art baselines, where it is typically one to two orders of magnitude faster to train. %;
%it also demonstrated some combinatorial generalisation properties.

As future work, we plan to apply it to broader RL domains (e.g., inducing game rules from a game trace) but also to continual learning scenarios where an interpretable logic-oriented higher-level policy would be particularly pertinent (e.g., autonomous driving). Indeed, we postulate that our model may be suited for continual learning in semantically-richer domains, for multiple reasons detailed in Appendix
~\ref{sec:promise}, alongside current limitations of our approach, which could be further investigated upon.

}

%\newpage
\bibliography{NSHRI}
\bibliographystyle{abbrvnat}

\newpage
\onecolumn
%%%%%%%%%%%%%%%%%%%%%%%%%%%%%%%%%%%%%%%%%%%%%%%%%%
\section*{Appendix}
%%%%%%%%%%%%%%%%%%%%%%%%%%%%%%%%%%%%%%%%%%%%%%%%%%
\setcounter{secnumdepth}{2}  
\renewcommand{\thesubsection}{\Alph{subsection}}
\renewcommand{\thesubsubsection}{\thesubsection.\arabic{subsubsection}}

\subsection{Expressivity Analysis}\label{appendix:meta_rules}

Many methods within inductive logic programming (ILP) define meta-rules, i.e.\pw{,} second-order Horn clauses which \pw{delineates} the structure of learnable programs. 
The choice of these meta-rules, referred to as a \textit{language bias} and restricting the hypothesis space, \pw{results} in a well-known trade-off between efficiency and expressivity.
In this section, we bring some theoretical justification for our designed set of second-order Horn Clause, by characterizing its expressivity.
Relying on minimal sets of metarules for rule induction models has been shown to improve both learning time and predictive accuracies \citep{cropper_logical_2014,fonseca_avoiding_2004}.
For a model to be both adaptive and efficient, it seems pertinent to aim for a minimal set of meta-rules generating a sufficiently expressive subset of Horn clauses.
The desired expressivity is ineluctably contingent to the tackled domain(s);
a commonly-considered fragment is Datalog $\mathcal{D}$ which has been proven expressive enough for many problems.
Datalog is a syntactic subset of Prolog which, by losing the Turing-completeness of Prolog/Horn, provides the undeniable advantage that its computations always terminate.
Below, we focus our attention to similar fragments, in concordance with previous literature \citep{galarraga_amie_2013, cropper_logical_2020}.

%We hope the reader to have a certain familiarity with logic programming notations, we quickly synthesize some core notations needed for our analysis.
%As we strongly follow the terminology employed by \citet{cropper_logical_2020}, we kindly suggest the reader to refer to it for further details.

\pw{To make the document self-contained, \cg{before stating our results, let us introduce both concepts and notations \pw{related to} First and Second Order Logic.}}
\pw{Recall $\Predicates$ denote the set of all considered predicates.}

\paragraph{Horn Logic}
\cg{We focus on Horn clause logic, since it} is a widely adopted\footnote{\pw{Pure} Prolog programs are composed by definite clauses and any \pw{query} in Prolog is a goal clause.} \textit{Turing-complete}\footnote{Horn clause logic and universal Turing machines are equivalent in terms of computational power.} subset of FOL. 
%if it can be used to simulate any Turing machine

A \textit{Horn clause} is a clause with at most one positive literal. \pw{Horn clauses \cg{may be of the following type\pw{s}}} \pw{(where $\mathsf{P}$, $\mathsf{Q}$, $\mathsf{T}$, and $\mathsf{U}$ are atoms)}:
\begin{itemize}
     \item \textit{goal clauses} which have no positive literal; they are expressed as $\lnot \mathsf{P} \lor \lnot \mathsf{Q} \lor ... \lor \lnot \mathsf{T}	$ or equivalently $\pw{\mathtt{False}} \leftarrow \mathsf{P} \land \mathsf{Q} \land ... \land \mathsf{T}$,
    \item \textit{definite clauses} which have exactly one positive literal; they are expressed as 
    	$\lnot \mathsf{P} \lor \lnot \mathsf{Q} \lor ... \lor \lnot \mathsf{T} \lor \mathsf{U}$ or equivalently \pw{as a \textit{Horn rule}} $\mathsf{U} \leftarrow \mathsf{P} \land \mathsf{Q} \land ... \land \mathsf{T}$,
     \item \textit{facts} which can be expressed as $\mathsf{U} \leftarrow \mathtt{True}$\footnote{Facts can be seen as a sub-case of definite clause assuming $\mathtt{True}\in \mathcal{P}$.} \pw{where $\mathsf{U}$ is grounded.}
\end{itemize}

\paragraph{Meta-rules}
We follow the terminology from the Meta Interpretative Learning literature \citep{cropper_logical_2020}:
\begin{definition}
A \emph{meta-rule} is a second-order Horn clause of the form: $A_0\leftarrow A_1\land... \land A_m$
where each $A_i$ is a literal of the form $P(T_1,...,T_{\pw{n_i}})$ where $P$ is \pw{a} predicate symbol or a second-order variable that can be substituted by a predicate symbol, and each $T_i$ is either a constant symbol or a first-order variable that can be substituted by a constant symbol.
\end{definition}
\pw{Here are s}ome examples of intuitive meta-rules, which \pw{will} be of later use:
\begin{equation}\label{eq:intuitive_metarules}
\begin{array}{lllll}
(\sigma) &\mathrm P(A, B)& \leftarrow& \mathrm Q(B, A)  & permute  \\
(\iota) &\mathrm P(A,B)&\leftarrow & \mathrm Q(A)   & expand  \\
(\exists)&\mathrm P(A) &\leftarrow &\exists B, \mathrm Q(A,B) & existential \:  contraction \\
(\forall)&\mathrm P(A) &\leftarrow &\forall B, \mathrm Q(A,B) & universal \: contraction \\
 (\nabla)&\mathrm P(A) &\leftarrow &\mathrm Q(A, A) & diagonal \:extract \\
(\Delta)&\mathrm P(A, A) &\leftarrow &\mathrm Q(A) & diagonal \:fill \\
\end{array}
\end{equation}
A common meta-rule set, which \pw{has} been used and tested in the literature \citep{cropper_logical_2020}, \pw{is the following}:
\begin{equation}\label{eq:cropper_metarules}
\PR_{MIL}=\left\lbrace \begin{array}{lllll}
(\texttt{Indent}_1) & &\mathrm P(A)& \leftarrow & \mathrm Q(A)    \\
(\texttt{DIndent}_1)  & &\mathrm P(A)& \leftarrow & \mathrm Q(A) \land \mathrm R(A)    \\
(\texttt{Indent}_2)  & &\mathrm P(A, B)& \leftarrow & \mathrm Q(A,B)    \\
(\texttt{DIndent}_2)  & &\mathrm P(A, B)& \leftarrow & \mathrm Q(A,B) \land \mathrm R(A,B)    \\
(\texttt{Precon})  & &\mathrm P(A, B)& \leftarrow & \mathrm Q(A) \land \mathrm R(A,B)    \\
(\texttt{Postcon})  & &\mathrm P(A, B)& \leftarrow & \mathrm Q(A,B) \land \mathrm R(B)    \\
(\texttt{Curry})  & &\mathrm P(A, B)& \leftarrow & \mathrm Q(A,B,\mathrm R)    \\
(\texttt{Chain})  & &\mathrm P(A, B)& \leftarrow &\mathrm Q(A,C) \land \mathrm R(C,B)   \\
\end{array}\right\rbrace
\end{equation}
\pw{where the letters $\mathrm P, \mathrm Q, \mathrm R$ correspond to existentially-quantified second-order variables and the letters $A, B, C$ to universally-quantified first-order variables.}

\paragraph{Proto-rules}
Our model \pw{relies} on an extended notion of meta-rules, which we refer to as \textit{proto-rules}.
These templates have the specificity that they do not fix the arity of their body predicates; only the arity of the head predicate is determined.
\pw{Formally, they can be defined as follows.
For any $n \in \mathbb N$, let $\Predicates^{n}$ (resp. $\Predicates^{\le n}$) be the predicates in $\Predicates$ that have arity equal to (resp. lower or equal to) $n$.

Define} the projection operator $\nu^{n}$ which canonically embeds predicates of arity $i\leq n$ in the space of predicates of arity $n$:
\begin{equation}\label{eq:projection}
\nu^{n}:\mathcal{P}^{\leq n} \longrightarrow \mathcal{P}^{n} \quad  \text{ s.t. } \quad 
\nu^{n}(P)(X_{1},\cdots, X_{n}) = P(X_{1},\cdots, X_{i}) \text{ for } P \in \mathcal{P}^{i}.
\end{equation}
Note that the restriction of $\nu^{n}$ on the space of predicates of arity n is identity: $\nu^{n} \mid_{\mathcal{P}^{n}} = id_{\mathcal{P}^{n}}$.
\pw{Moreover, this projection naturally extends to second-order variables.}
To ease the notations, we denote below the projection $\nu^{n_i}$ with an overline (e.g., for a predicate $P$, $\nu^{n_i}(P) = \overline P$), since there is no risk of confusion because $n_i$ is specified by its arguments $(T_1, \ldots, T_{n_i})$.

\begin{definition}
A \emph{proto-rule} is an extension of a second-order Horn clause of the form: 
%$A_0\leftarrow \overline{P}(T_1,...,T_{n_{1}}) \land... \land  \overline{P}(T_1,...,T_{n_{m}})$,
% This notation seems to suggest that each predicate has the same initial terms.
\pw{$A_0\leftarrow A_1 \land \ldots \land A_m$,}
where  $A_0$ \pw{(resp. each $A_i, i>0$} is \pw{a literal of the form $P(T_1,...,T_{n_{i}})$ \pw{(resp. $\nu^{n_i}\big({P}\big)(T_1,...,T_{n_{i}})$} in which $P$\pw{, which} is a} predicate symbol or a second-order variable that can be substituted by a predicate symbol, \pw{is of arity lower or equal to $n_i$}, and each $T_i$ is either a constant symbol or a first-order variable that can be substituted by a constant symbol.
\end{definition}
\cg{Seeing second-order rules as functions over predicate spaces to first-order logic rules, we can provide another characterization of proto-rules:}
\pw{a meta-rule can be understood as a mapping \cg{$\Predicates^{n_1} \times \ldots \times \Predicates^{n_m} \rightarrow \mathcal{H}$ for some indices $n_i$}, i.e., taking $m$ predicates of specific arities and returning a first-order \cg{Horn clause};
in contrast, a proto-rule is a mapping $\Predicates^{\le n_1} \times \ldots \times \Predicates^{\le n_m} \rightarrow \mathcal{H}$, for some indices $n_i$.
}

The first set of protorules we \pw{propose} is the following:
\begin{equation}\label{eq:protorule_set}
\PR_0:=\left\lbrace \begin{array}{llll}
 \mathfrak{A}: & P(A) &\leftarrow &  \overline{Q}(A,B) \land \overline{R}(B,A) \\
 \mathfrak{B}:  & P(A,B) &\leftarrow &  \overline{Q}(A,C) \land \overline{R}(C,B) \\
\mathfrak{C}:  & P(A,B) &\leftarrow &  \overline{Q}(A,B) \land \overline{R}(B,A) \\
\end{array}
\right\rbrace
\end{equation}

\paragraph{Horn Fragments}

A \textit{Horn theory} $\mathcal{T}$ is a set of Horn clauses. 

\cg{
Let us denote the set of second order Horn clause\pw{, the set of} definite Horn clauses\pw{, and the set of} function-free definite Horn clauses \pw{respectively} by $\mathcal{H}$, $\mathcal{S}$, and $\mathcal{F}$.
%Let us point out that these Horn theories are implicitly depending on the predicate set $\Predicates$, and should be written as 
%$\mathcal{H}_\Predicates$, resp. $\mathcal{S}_\Predicates$ resp. $\mathcal{F}_\Predicates$;
%however, for the sake of simplicity,
%we omit this dependency in our notations \footnote{It is common, in some FOL literature, to consider the predicate set infinite, in order to consider only one language of first-order logic.}.
}
%All the clauses considered in \pw{the remaining are assumed to be} function-free definite second order Horn clauses, i.e., in $\mathcal{F}$, although it could be transposed in a broader context.
Within Horn clause logic, several restrictions have been proposed in the literature to narrow the hypothesis space, \cg{under the name of \emph{fragment}.
A \emph{fragment} is a syntactically restricted subset of a theory.}
\cg{Following previous work (such as \citep{cropper_logical_2020}), we introduce below a list of classic fragments of $\mathcal{F}$, in decreasing order:}

\begin{itemize}
    \item \textit{connected} fragment, $\mathcal{C}$: connected definite function-free clauses, i.e., clause\pw{s} in $\mathcal{F}$ whose literals can not be partitioned into two sets such that the variables attached to one set are disjoint from the variables attached to literal\pw{s} in the other set;
    e.g., $P(A,B) \leftarrow Q(A,C) \land R(A,B)$ is connected while the following clause is not $P(A,B) \leftarrow Q(A,C) \land R(B)$.
    \item \textit{Datalog}\footnote{Some wider Datalog fragment have been proposed, notably allowing negation in the body, under certain stratification constraints; here we consider only its intersection with function-free definite Horn clauses.} fragment, $\mathcal{D}$: sub-fragment of $\mathcal{C}$, composed of clauses such that \cg{every variable present in the head of the Horn rule also appears in its body.}
    \footnote{Since below each clause is assume \pw{to be} a definite Horn clause, it can be equivalently represented as a Horn rule, with one head and several non-negated literals in the body; we below alternately switch between these representations.}
    \item \textit{two-connected} fragment, $\mathcal{K}$: sub-fragment of $\mathcal{D}$ composed of clauses such that the variables appearing in the body must appear at least twice.
    \item \textit{exactly-two-connected} fragment, $\mathcal{E}$: sub-fragment of $\mathcal{K}$ such that the variables appearing in the body must appear exactly twice.
    \item \textit{duplicate-free} fragment, $\mathcal{U}$: sub-fragment of $\mathcal{K}$ excluding clauses in which a literal contains multiple occurrences of the same variable, e.g., excluding $P(A,A) \leftarrow Q(A)$.
\end{itemize}
%The different fragments mentioned above \pw{are related as follows}:
%\begin{equation}
  %\mathcal{H} \supset     \mathcal{S} \supset \mathcal{F} \supset	 \mathcal{C}\supset	  \mathcal{D}\supset	  \mathcal{K} \begin{array}{c}
  %       \rotatebox[origin=c]{25}{$\supset$} \hspace{.5em} \mathcal{E} \\
  %       \rotatebox[origin=c]{-25}{$\supset$} \hspace{.5em} \mathcal{U}
  %   \end{array} 
%\end{equation}

These fragments  are related as follows:
\begin{equation}
  \mathcal{H} \supset    \mathcal{S} \supset\mathcal{F} \supset	 \mathcal{C}\supset	 \mathcal{D}\supset	  \mathcal{K} \begin{array}{c}
         \rotatebox[origin=c]{25}{$\supset$} \hspace{.5em} \mathcal{E} \\
         \rotatebox[origin=c]{-25}{$\supset$} \hspace{.5em} \mathcal{U}
     \end{array} 
\end{equation}

For $\mathcal{M}$ a set of meta-rules, we denote $\mathcal{M}^{\leq a}_{\leq m}$ the fragment of $\mathcal{M}$ where each literal has arity at most $a$ and each \pw{meta-rule} has at most $m$ body literals.
We extend the notation in order to allow both $a$ and $m$ to be a set of integers; e.g., $\mathcal{M}^{\lbrace 1,2\rbrace} _{\lbrace 3\rbrace }$ is the fragment of $\mathcal{M}$ where each literal \pw{corresponds to} an unary or binary predicate and \pw{the body of} each \pw{meta-rule} contains exactly three literals.
%The rule space enabled by $\mathcal{M}^{a}_{m}$ is bounded bz....

\paragraph{Binary Resolution}
%cf GOOD REFERENCE: http://www.cs.ru.nl/~peterl/teaching/KeR/logicintro.pdf

As it would be of future use, let us introduce some notions around resolution.
To gain intuition about the notion of \textit{resolvent}\footnote{Another way to gain intuition, is to think about propositional logic; 
there, a resolvent of two parent clauses containing complementary literals, such as $P$ and $\lnot P$, is simply obtained by taking the disjunction of these clauses after removing these complementary literals.
In the case of FOL or HOL, such resolvent may involve substitutions.
},
let us look at an example before stating more formal definitions. Consider the following Horn clauses:
$$
\begin{array}{llll}
C_1: & P(X,Y) \lor \lnot Q(X) \lor  \lnot R(X,Y) & \text{ or, equivalently, }&  P(X,Y) \leftarrow Q(X) \land R(X,Y) \\
C_2:  & R(X,a) \lor \lnot T(X,a)  \lor \lnot  S(a) & \text{ or, equivalently, } & R(X,a) \leftarrow S(a) \land T(X,a)
\end{array}$$
The atoms $R(X,Y)$ and $R(X,a)$ are unifiable, under the substitution $\sigma:\lbrace a/Y \rbrace $;
under this substitution, we obtain the following clauses:
$$\lbrace P(X,a) \lor \lnot Q(X) \lor  \lnot R(X,a), R(X,a) \lor \lnot T(X,a)  \lor \lnot  S(a) \rbrace $$
Since either $ R(X,a)$ is True, or $ \lnot R(X,a)$ is True, assuming $C_1$ and $\pw{C_2}$ are True, we can deduce the following clause is True:
$$(P(X,a) \lor \lnot Q(X)) \lor ( \lnot T(X,a)  \lor \lnot  S(a) )  \text{ or, equivalently, } P(X,a) \leftarrow Q(X) \land T(X,a)  \land  S(a) $$
This resulting clause, is a binary resolvent of $C_1$ and $C_2$.

Now, let us state more formal definitions. We denote $\sigma = \lbrace t_1/\pw{X}_1, \cdots , t_n/\pw{X}_n\rbrace,$ a substitution and $E\sigma$ the expression obtained from an expression $E$ by simultaneously replacing all occurrences of the variables $\pw{X}_i$ by the terms $t_i$.

\begin{definition}\label{unifier}
\begin{itemize}
    \item  A substitution $\sigma$ is called a unifier of a given set of expressions $\lbrace E_1, \cdots , E_m\rbrace$ , $m \geq 2$ if $E_1\sigma = \cdots  = E_m\sigma$.
    \item  A unifier $\theta$ of a unifiable set of expressions $ E = \lbrace E_1,\cdots  , E_m\rbrace$ , $m \geq 2$, is
said to be a most general unifier if for each unifier $\sigma$ of E there exists a substitution $\lambda$
such that $\sigma = \theta\lambda$.
\end{itemize}
\end{definition}

\begin{definition}\label{binary_resolvent_proposition}
Consider two parent clauses $C_1$ \pw{and} $C_2$ containing the literals $l_1$ \pw{and} $l_2$ \pw{respectively}. 
If $l_1$ and $\lnot l_2$ have a most general unifier $\sigma$, then the clause $(C_1\sigma\setminus l_2\sigma) \lor (C_2\sigma\setminus	l_2\sigma)$ is called a binary resolvent of $C_1$ and $C_2$.
%Resolution in which each resolvent is a binary resolvent, is known as binary resolution.
\end{definition}

\paragraph{Derivation Reduction}
Diverse reductions methods have been proposed in the literature, such as  subsumption \citep{robinson_machine_65}, entailment \citep{Muggleton2009InverseEA}, and derivation \citep{cropper_derivation_2018}. 
As previously pointed out (e.g., \citep{cropper_logical_2020}), common entailment methods may be too strong and remove useful clause\pw{s} enabling to make predicates more specific.
The relevant notion to ensure the reduction would not affect the span of our hypothesis space is \pw{the} one of derivation-reduction (\textit{D-reduction}) \citep{cropper_derivation_2018}, as defined in Definition~\ref{derivation}.

Let us first define for the following operator \pw{for a Horn theory $\mathcal T$}:
$$\left\lbrace \begin{array}{lll}
   R_0(\mathcal{T}) &=& \mathcal{T}   \\
   R_n(\mathcal{T})  & = & \lbrace C\mid C \text{ is the binary resolvent of } C_1 \text{ and } C_2 \text{ with } C_1 \in R_{n-1}(\mathcal{T}),C_2 \in \mathcal{T}  \rbrace
\end{array}\right.$$

\begin{definition}\label{horn_clausure}
The \emph{Horn closure} of a Horn theory $\mathcal{T}$ is: $\displaystyle R^{\ast}(\mathcal{T} )= \cup_{n} R_n(\mathcal{T})$.
\end{definition}
\begin{definition}\label{derivation}
\begin{itemize}
    \item[$(i)$] A Horn clause $C$ is \emph{derivationally redundant} in the Horn theory $\mathcal{T}$ if $\mathcal{T} \vdash C$, i.e., $C \in R^{\ast}(\mathcal{T})$; it is said \pw{to be} \emph{$k$-derivable} from $\mathcal{T}$ if $C \in R^{k}(\mathcal{T})$.
     \item[$(ii)$] A Horn theory $\mathcal{T}$ is derivationally reduced \pw{($D$-reduced)} if and only if it does not contain any derivationally redundant clauses.
\end{itemize}
\end{definition}

%"(D-reduction problem Horn decidability) The D-reduction problem for Horn theories is undecidable. However, the D-reduction problem is decidable for any fragment Mma (e.g., definite Datalog clauses where each clause has at least one body literal, with additional arity and body size constraints). 
%cf Cropper and Touret 2020"

A clausal theory may have several D-reductions. For instance, we can easily see that
$$\left\lbrace\begin{array}{llll}
C_1:& P(A, B)& \leftarrow& Q(B, A)    \\
C_2:& P(A,B)&\leftarrow &Q(A,C),R(C,B)     \\
C_3:& P(A, B) &\leftarrow &Q(A, C), R(B, C)
\end{array} \right\rbrace \text{ may be D-reduced to either } \lbrace C_1,C_2 \rbrace  \text{ or }  \lbrace C_1, C_3 \rbrace $$

\paragraph{Hypothesis Space}
\pw{Recall} that $\Predicates$ \pw{(resp. $\Predicates_{0}$) denotes} the set of predicates \pw{(resp. of initial predicates)};
and $\mathcal{B}$ the background knowledge, may encompass both initial predicates and pre-given rules.
\cg{
As we are interested \pw{in} characterizing the expressivity of our model, let us define the relevant notion of hypothesis space\footnote{In this paper, we adopted the denomination of \emph{hypothesis space}, \pw{which is unconventional in logic, because} it refers to the space accessible to our learning algorithm.
Similarly, by incorporating initial rules, we could also define the Horn theory generated by a set of meta-rules or proto-rules given a certain background knowledge $\mathcal{B}$ \pw{and a set of predicates $\mathcal P$.}}, which could apply to any rule-induction model attached to at meta-rule or proto-rule set:
\begin{definition}\label{hypothesis_space_definition}
Given a predicate set $\Predicates$, and a meta-rule or proto-rule set $\mathcal{M}$, let us define:
\begin{itemize}
    \item[$(i)$] the set $\mathcal{M}_{\Predicates}$ as the set of all Horn clauses generated by all the possible substitutions of predicates variables in $\mathcal{M}$ by predicates symbols from $\Predicates$.
    \item[$(ii)$]  the \emph{hypothesis space} $\mathcal{M}[\Predicates]$ generated by $\mathcal{M}$ from $\Predicates$ as the Horn closure of $\mathcal{M}_{\Predicates}$.
\end{itemize}
\end{definition}
}
\paragraph{Equality Assumption}
\cg{The Equality relation is typically assumed part of FOL, so it is \pw{reasonable} to assume it belongs to the background knowledge in our results below.
In our approach, it amounts to integrat\pw{ing} the predicate $\mathtt{Equal}$ to the set of initial predicates \pw{$\Predicates_0$};
$\mathtt{Equal}$, which corresponds to the identity, is inten\pw{s}ionally defined on any domain $\mathbb{D}$ by:

}
\begin{equation}\label{eq:same}
\mathtt{Equal}:\left\{
\begin{array}{lll}
\mathbb{D}\times  \mathbb{D}&\rightarrow & \pw{\mathbb B} = \lbrace    \mathtt{True} ,   \mathtt{False}  \rbrace \\
(X,Y)& \mapsto & \left\lbrace
\begin{array}{ll}
\mathtt{True}  & \text{ if } X=Y   \\
\mathtt{False}  &\text{ if }  X\neq Y  
\end{array} \right.
\end{array}\right.
\end{equation}

\paragraph{Main Result}
Here \pw{is} our main result, concerning the investigation of the expressivity of the minimal set $\PR_0$:

\begin{theorem}\label{thm:expressivity}
\begin{enumerate}
    \item[$(i)$] Assuming $ \mathtt{True} \in \Predicates_0$, the hypothesis space generated by $\PR_0$ from $\Predicates$ encompasses the set of duplicate-free and function-free definite Horn clause composed of \pw{clauses} with \pw{at most} two body atoms involving unary and binary predicates in $\Predicates$:
    \pw{$$\mathtt{True} \in \Predicates_0 \implies  \PR_0[\Predicates] \supset \mathcal U_{\leq 2}^{\lbrace 1,2 \rbrace}[\Predicates]$$}
    \item[$(ii)$] Assuming $ \mathtt{True}, \mathtt{Equal} \in \Predicates_0$, the hypothesis space generated by $\PR_0$ from $\Predicates$ corresponds to the set of function-free definite Horn clause composed of clauses with \pw{at most} two body atoms involving unary and binary predicates in $\Predicates$;
    i.e., in terms of the second order logic fragment:
    $$\mathtt{True},\mathtt{Equal}  \in \Predicates_0 \implies  \PR_0[\Predicates] = \mathcal F_{\leq 2}^{\lbrace 1,2 \rbrace}[\Predicates]$$
\end{enumerate}
\end{theorem}
We refer to the end of this section for the proof of this theorem.

% Let us point out that other alternative proto-rules set, equally expressive\footnote{Indeed, the proto-rule $ \mathfrak{A}$ can be written as composition $(\nabla)\circ ( \mathfrak{B})$ or $(\exists)\circ ( \mathfrak{C})$. Reciprocically the proto-rule $(\exists)$ can be written as $\mathfrak{A}(\cdot, \mathtt{True})$, which leads to the claim.}, could be designed, such as:
% \begin{equation}
% \PR_{\exists}:= \left\lbrace \exists,  \mathfrak{B},  \mathfrak{C}\right\rbrace
% \end{equation}
%This would be \pw{the} object of further investigation and experiments.
% I think it's better not to mention that for now, if we plan to do it.

\pw{Theorem~\ref{thm:expressivity}} allows us to \pw{conclude that} $\PR_0$ is more expressive than the set commonly found in the literature $\PR_{MIL}$, assuming we are working with predicates of arity at most $2$:
\begin{corollary}
Assuming the initial predicates $\mathcal{P}_{0}$ contains only predicates of arity at most $2$, the hypothesis space generated by $\PR_0$ given $\Predicates$ encompass\pw{es the} one generated by $\PR_{MIL}$ as defined in (\ref{eq:cropper_metarules}).
$$ \forall P\in \mathcal{P}_{0},\quad  \mathrm{arity}(P) \leq 2 \implies \PR_0[\Predicates]\supset \PR_{MIL}[\Predicates]$$
\end{corollary}
\begin{proof}
Under our assumption, the meta-rule $(\texttt{Curry})$ in (\ref{eq:cropper_metarules}) may be disregarded. 
The remaining \pw{meta-}rules present in $\PR_{MIL}$ have already been examined in Theorem~\ref{thm:expressivity}. $\PR_0$ has therefore at least the same expressivity than $\PR_{MIL}$. In order to be able to conclude $\PR_0$ is strictly more expressive, we can mention the rules $P(A) \leftarrow P(A,B)$ or $P(A, B)\leftarrow P(B,A)$, which are reached by $\PR_0$ not $\PR_MIL$.
\end{proof}

\paragraph{Disjunction}
In our model, to ensure an incremental learning of the target rule, we impose the restriction that each auxiliary predicate corresponds to one rule. 
Since such constraint risk to remove the possibility of disjunction, we redefine new proto-rules versions integrating disjunctions:
\begin{equation}\label{eq:protorule_or}
\PR_0^\lor=\left\lbrace \begin{array}{llll}
 \mathfrak{A}: & P(A) &\leftarrow & \left( \overline{Q}(A,B) \land \overline{R}(B,A) \right)\lor \overline{S}(A,B) \\
 \mathfrak{B}:  & P(A,B) &\leftarrow &\left(  \overline{Q}(A,C) \land \overline{R}(C,B)\right) \lor \overline{S}(A,B)\\
\mathfrak{C}:  & P(A,B) &\leftarrow &  \left(\overline{Q}(A,B) \land \overline{R}(B,A) \right) \lor \overline{S}(A,B)\\
\end{array}
\right\rbrace
\end{equation}

Adopting the minimal set $\PR_0^\lor$, in our incremental setting does not affect the expressivity of the hypothesis space compared to when relying on the set $\PR_0$ in a non incremental setting.
Let us remind an \emph{incremental setting} imposes a predicate symbol to be the head of exactly one rule.

\begin{lemma}
The minimal proto-rule set $\PR_0^\lor$ in an \emph{incremental setting} holds the same expressivity \pw{as} $\PR_0$.
\end{lemma}
\begin{proof}
It is straightforward to see that $\PR_0^\lor$ is a minimal set.\\
The hypothesis space generated by $\PR_0$ obviously encompasses the hypothesis space generated by $\PR_0^\lor$ in an incremental setting. Reciprocally, let us assume $n$ rules instantiated from $\PR_0$ are attached to one predicate symbol $P$. They can be expressed as:
$$
\begin{array}{ccc}
    P(X,Y)  & \leftarrow & f_1(X,Y)  \\
    P(X,Y)  & \leftarrow & f_2(X,Y)  \\
     \cdots &  \cdots &  \cdots \\
       P(X,Y)  & \leftarrow & f_n(X,Y) \\
\end{array}
$$
Recursively, we can define $P_{i}$ auxiliary predicates in the hypothesis space of $\PR_0^\lor$.
$$
\begin{array}{ccc}
    P_1(X,Y)  & \leftarrow & f_1(X,Y)\\
    P_2(X,Y)  & \leftarrow &  P_1(X,Y) \lor  f_2(X,Y) \\
     \cdots &  \cdots &  \cdots \\
       P_{n}(X,Y)  & \leftarrow &  P_{n-1}(X,Y)\lor  f_n(X,Y) \\
\end{array}
$$
Since $  P_{n}$ is then equivalent to $P$, we can conclude.

\end{proof}

\paragraph{Redundancy}
Since we have observed a certain redundancy to be beneficial to the learning, we incorporate the permutation rule $\mathfrak{I}$ in our experiments, and rely on the following set:
\begin{equation*}
\PR_*=\PR_0^\lor \cup  \lbrace \mathfrak{I}:  H(X, Y)  \leftarrow \overline{F}(Y, X) \rbrace
\end{equation*}
This small---yet non minimal---set of protorules \pw{$\PR_*$}, has the same expressivity as $\PR_0$ (\pw{characterized \cg{in Theorem \ref{thm:expressivity}})} 

\begin{lemma}
The proto-rule set $\PR_\ast$ holds the same expressivity than $\PR_0$.
\end{lemma}
\begin{proof}
It stems from the fact, observed in the proof of Theorem \ref{thm:expressivity} that the rule $(\sigma)$, equivalent to $(\mathfrak{I})$ can be D-reduced from $\PR_0$.
\end{proof}

\paragraph{Recursivity}
The development of Recursive Theory arose first around the Hilbert’s Program and G\"odel’s proof of the Incompleteness Theorems (1931) and is tightly linked to questions of computability.
While implementing recursive-friendly model, we should \pw{keep} in mind considerations of computability and decidability. 
Enabling full recursivity within the templates is likely to substantially affect both the learning and the convergence, by letting room to numerous inconsistencies.\\

With this in mind, let us consider different stratifications in the rule space, and in our model. First, we introduce a \textit{hierarchical filtration} of the rule space: initial predicates are set of \pw{layer} 0, and auxiliary predicates in \pw{layer $\ell$} are defined from predicates from \pw{layers at most $\ell$}. Within such stratified space, different restrictions may be \pw{considered:} %grafted touching upon the recursivity.\\
% STRATIFICTAION of logic program ?? \citep{sebellk_horn_1982}
%For instance, let $\mathcal{F}_{d}$ the depth filtration attached to our template set $\PR^{\lor}_{0}$ such that: the level 0 are the initial predicates $\mathcal{F}_{0}=\mathcal{P}_{0}$ and $\mathcal{F}_{d}$ the predicates which may be intensionally defined through one template $\tau\in\PR^{\lor}_{0}$ involving only body predicates of smaller level.

\begin{itemize}
\item \textit{recursive-free hierarchical hypothesis space}:, where rules of \pw{layer $\ell$} are defined from rules of strictly \pw{lower layer}. 
The \pw{proto-rules from $\PR_0^\lor$} are therefore restricted to the following form:
$$ P(\cdot) \leftarrow  \left[ Q(\cdot) \land R(\cdot) \right] \lor  O(\cdot),$$
with $P\in \mathcal{P}_{\ell}$, $Q,R, O\in \mathcal{P}_{< \ell}$.
\item  \textit{iso-recursive hierarchical hypothesis space}, where the only recursive rules allowed \pw{from $\PR_0^\lor$} have the form:
$$ P(\cdot) \leftarrow  \left[ Q(\cdot) \land R(\cdot) \right] \lor  O(\cdot), \quad \text{ with } P\in \mathcal{P}_{\ell}, Q,R\in \mathcal{P}_{<\ell}\cup \lbrace P \rbrace , O\in \mathcal{P}_{<\ell}$$
Such space \pw{fits} most recursive ILP tasks (e.g., Fizz, Buzz, Even, LessThan). 
\item  \textit{recursive hierarchical hypothesis space}, where the recursive rules allowed \pw{from $\PR_0^\lor$} have the form:
$$ P(\cdot) \leftarrow  \left[ Q(\cdot) \land R(\cdot) \right] \lor  O(\cdot) \quad  \text{ with } P\in \mathcal{P}_{\ell}, Q, R\in \mathcal{P}_{\le\ell}, O\in \mathcal{P}_{\le\ell}.$$
Such space is \pw{required} for the recursivity present in EvenOdd or Length. 
Such hierarchical notion is also present in the Logic Programming literature under the name of \emph{stratified programs}.
\end{itemize}

%Common classes of computable recursive functions are \textit{primitive recursive} \textit{PrimRec} and \textit{partial recursive functions} \textit{PartRec}.
%For instance, a \textit{PrimRec} may be defined as the closure under the operators $\lbrace \textit{Comp}, \mbox{\textit{PrimRec}} \rbrace$ (namely composition and primitive recursion) of the set of initial functions $I=\lbrace zero, succ, \pi_{i}^{k}\rbrace$ where $ \pi_{i}^{k}$ are canonical projections functions on each variable. To extend this class to Partially Recursive functions, we complement with an extra operator Min, of Minimization. 

%Although defined initially for multi-variate functions on natural numbers, these definitions may be extended to any alphabet  but also have been applied to predicates defined on such an alphabet, through their characteristic functions. \footnote{As an n-ary predicate P (over $\Sigma$) is any subset of $(\Sigma^{\ast})^{n}$, which can be rewritten as an n-ary function on $\Sigma^{\ast}$.}
%(Note that both the arithmetic or List domains may be reframed as such examples.)

\paragraph{Proof of Theorem \ref{thm:expressivity}}
\cg{Before \pw{we present the proof}, let us introduce two \pw{additional} notations: 
\begin{itemize}
    \item $\mathfrak{R} \prescript{}{i}{\odot} \mathfrak{R}' $ \pw{refers} to the resolvent of $\mathfrak{R}$ with $\mathfrak{R'}$, where the $i^{th}$ body member of $\mathfrak{R}$ is resolved with the head of $\mathfrak{R}'$. 
    To illustrate it, consider the following meta-rules:
$$\begin{array}{lc}
\pw{\mathfrak{R}}: & P(A) \leftarrow Q(A) \land R(A,B)\\
\pw{\mathfrak{R}'}: & P(A) \leftarrow Q(A) \land R(B,A)\\
\pw{\mathfrak{R}''}: & P(A,B) \leftarrow Q(B,A)
\end{array}
$$
The resolvent $\mathfrak{R}' \prescript{}{2}{\odot} \pw{\mathfrak{R}''} $ corresponds to resolve the second body atom of $\pw{\mathfrak{R}'}$ with $\pw{\mathfrak{R}''}$, \pw{which therefore leads to} $\pw{\mathfrak{R}}$.
\item We define the operation $\rho$ as the composition of a \pw{projection} $\nu$, a permutation $(\sigma)$, and an existential contraction over the second variable $(\exists)$, which amounts to:
\begin{equation}\label{eq:rho}
\rho(R)(X):=\exists \circ \sigma (\overline{R}(X,Y)) = \exists Y R(Y)
\end{equation}
Here, we identified the permutation clause $(\sigma)$ (as defined in (\ref{eq:intuitive_metarules})) with the operation it defines on predicates $\sigma$; similarly for the existential clause $(\exists)$ with the contraction operation $\exists$.

Likewise, we can define the corresponding non-connected meta-rule attached to the operation $\rho$ by:
\begin{equation}\label{eq:rho2}
(\rho):\: P(X) \leftarrow  \exists Y R(Y)
\end{equation}
\end{itemize}
}
\begin{proof}
The inclusion $\PR_0[\Predicates] \subset  \mathcal{F}_{\leq 2}^{\lbrace 1,2 \rbrace}[\Predicates]$ is trivial since all the rules instantiated \pw{from} $\PR_0$ are definite Horn clause with two body predicates, and duplicate-free; similarly, if adding the \pw{duplicate-free constraint} to $\mathcal{F}$ (and denoting \pw{it} by $\mathcal{F}^{D}$), we can trivially state: $\PR_0[\Predicates] \subset  \mathcal{F}_{\leq 2}^{D, \lbrace 1,2 \rbrace} [\Predicates]$.

It remains to prove the other inclusion; e.g., for $(ii)$, it amounts to $\PR_0[\Predicates] \supset  \mathcal{F}_{\leq 2}^{\lbrace 1,2 \rbrace}[\Predicates]$, which boils down to $\PR_0[\Predicates] \supset  \mathcal{F}_{\Predicates, \leq 2}^{\lbrace 1,2 \rbrace}$ since  $\PR_0[\Predicates]$ is closed under resolution.
We will proceed by successively extending the sub-fragments $\mathcal{M}$ we are considering while proving $\PR_0[\Predicates] \supset \mathcal{M}_{\Predicates} $.
For instance, under the assumption $\mathtt{True}, \mathtt{Equal}\in \Predicates_0$, we demonstrate that the hypothesis space generated by $\PR_0$ encompasses the one generated by the increasingly \pw{larger} fragments:
(a) $\mathcal{U}$ (b) $\mathcal{K}$; (c) $\mathcal{D}$; (d) $\mathcal{C}$; (e) $\mathcal{F}$. For $(i)$, under the simpler assumption $\mathtt{True} \in \Predicates_0$, we can follow almost identical steps, although the duplicate-free constraint can not be lifted, and therefore step $(b)$ is skipped. Since the proof of $(i)$ is more or less included in the proof of $(ii)$, we will below focus only on proving $(ii)$.

For each fragment $\mathcal{M}$, to demonstrate such inclusion, we can simply show that every rule within the fragment $\mathcal{M}_{\Predicates}$ belongs to the hypothesis space generated by $\PR_0$;
more specifically, we will mostly work at the level of second order logic; there, we will show that any meta-rule in $\mathcal{M}$ can be reduced from meta-rules in $\PR_0$, or generated by $\PR_0$.

However, to avoid listing all the meta-rules generating these fragments, and restrict our enumeration, we leverage several observations: 
first, since we assume the predicate set includes $\mathtt{True}$, we can restrict our attention to clauses with exactly two body literals;
secondly, the permutation meta-rule $(\sigma)$ (introduced in \ref{eq:intuitive_metarules}) can be derived from the proto-rule $\mathfrak{C}$ upon matching the first body in $\mathfrak{C}$ with $\mathtt{True}$;
therefore, we can examine rules only \cg{upon} permutation $(\sigma)$ applied to their body or head predicates.
%finally, since conjunction is a symmetric operation, we can consider only meta-rules \cg{upon} permutation of their body predicates.

\begin{enumerate}
    \item First, let us demonstrate the following inclusion:
 \begin{equation}\label{eq:u}
  \mathtt{Claim}_1:  \quad\mathtt{True}  \in \Predicates_0\implies \PR_0[\Predicates]  \supset \mathcal{U}_{\Predicates, \leq 2}^{\lbrace 1,2 \rbrace}
  \end{equation}
By the previous observations, we can narrow down to examine a subset of the meta-rules generating $\mathcal{U}_{\leq 2}^{\lbrace 1,2 \rbrace}$, as listed and justified below.
Our claim is that all these clauses are derivationally redundant from the Horn theory generated by $\PR_0$;
for each of them, we therefore explicit the clauses used to reduce them\footnote{Note that we are allowed to use clauses $\exists$ and $\sigma$ in the reduction as we already explained why we can derive them from $\PR_0$ in the beginning of the proof of Theorem \ref{thm:expressivity}.
}.\\
$$
\begin{array}{clll}
 &\text{ Meta-Rules }& \text{Reduction}  & \text{Comment} \\
   (i) & P(A) \leftarrow  Q(A)\land R(A) &  \mathfrak{A} 
   \prescript{}{2}{\odot} \sigma \quad & \mathfrak{A}(Q,  \sigma(R)) \\
   (ii)&   P(A)  \leftarrow  Q(A)  \land R(A,B) &  \mathfrak{A} 
   \prescript{}{2}{\odot} \sigma  & \mathfrak{A}(Q,  \sigma(R)) \\
    (iii)&  P(A)  \leftarrow  Q(\pw{A,} B) \land R(\pw{B}) &  \mathfrak{A}  & \\
    (iv) & P(A)  \leftarrow  Q(A,B) \land R(B,A) &  \mathfrak{A}& \\
  (v)&    P(A)  \leftarrow   Q(A,B)  \land R(B,C) &\mathfrak{A}&  \\
    (vi)&   P(A,B)  \leftarrow   Q(A)  \land R(A, B) & \mathfrak{A} 
   \prescript{}{2}{\odot}\sigma & \mathfrak{C}(Q,  \sigma(R))\\
    (vii) & P(A,B)  \leftarrow  Q(A,C) \land  R(C,B)& \mathfrak{B}&  \\
     (viii)&  P(A,B)  \leftarrow  Q(A,B) \land R(B,A) & \mathfrak{C}&\\
     (ix) & P(A,B)  \leftarrow  Q(A,B) \land  R(B,C)& \mathfrak{A} 
   \prescript{}{2}{\odot}\exists& \mathfrak{A}(Q,  \exists(R)) \\
\end{array}
$$
Beside the notation of the resolvent $\odot_H$ explained previously, we provided a more intuitive form for the resolution (under 'Comment'), in terms of composition between maps, where we identify a proto-rule (or meta-rule) with a map of the following form:
$$\mathcal{P}^{\bullet} \times \cdots \times \mathcal{P}^{\bullet} \rightarrow \mathcal{P}^{\bullet}$$
 \cg{Let us justify why these clauses are the only clauses in $\mathcal{U}_{\leq 2}^{\lbrace 1,2 \rbrace}$ worth to examine in two steps:
\begin{itemize}
    \item Let us consider clauses whose head predicate arity \pw{is} $1$.
First, by the Datalog assumption, we can indeed restrict our attention to body clauses where $A$ is appearing at least once. 
By symmetry, we can assume $A$ appears in the first body. 
By the connected assumption, either $A$ appears in the second body too, and/or an existential variable (say $B$) appears both in the first or second body. 
Upon permutation, it leads us to clauses $(i)-(v)$.
 \item Let us consider clauses whose head predicate arity \pw{is} $2$.
First, by the Datalog assumption, we can indeed restrict our attention to body clauses where $A$ and $B$ have to appear at least once.
The only body clause of arity $(1,1)$ satisfying this condition is not connected ($ Q(A)\land R(B)$); 
upon permutation of $A$ and $B$, and of the body predicates $Q$ and $R$, and upon inversion $(\sigma)$ of the body predicates, the only body clause of arity $(1,2)$ is of the form $ Q(A) \land R(A, B)$ (denoted  $(vi)$ below, which reduces to $\mathfrak{C}$ upon permutation); 
for arity $(2,2)$, upon similar permutations, we can list three type\pw{s} of body clause, listed below as $(vii),(viii), (ix)$. 
While $(vii),(viii)$ are directly pointing to $\mathfrak{B}, \mathfrak{C}$, $(ix)$, can be obtained by reducing the second body of $ \mathfrak{C}$ through $(\exists)$ (as defined in \ref{eq:intuitive_metarules}).
This justifies the remaining clauses, $(vi)- (ix)$.
\end{itemize}
}
We can therefore conclude by $\mathtt{Claim}_1$.

\item  Let us extend \pw{(\ref{eq:u})} by enabling duplicates in the clause:
 \begin{equation}\label{eq:k}
 \mathtt{Claim_2:}\quad \mathtt{True},\mathtt{Equal}  \in \Predicates_0 \implies \PR_0[\Predicates] \supset \mathcal{K}_{\Predicates, \leq 2}^{\lbrace 1,2 \rbrace} 
   \end{equation}
To extend the result from $\mathcal{U}_{\leq 2}^{\lbrace 1,2 \rbrace}$ to $\mathcal{K}_{\leq 2}^{\lbrace 1,2 \rbrace}$, it is sufficient to show the  meta-rules $ (\nabla),(\Delta)$ \pw{from} (\ref{eq:intuitive_metarules}) can be derived from $\PR_0$, under the extra-assumption that $\mathtt{Equal}$ is included as background knowledge.
This stems from the fact that these meta-rules can be written as follows:
$$
\begin{array}{cll}
 &\text{ Meta-rule }& \text{Equivalent Form}\\
(\nabla) & P(A,A)  \leftarrow  Q(A)& P(A,B)  \leftarrow  Q(A) \land  \mathtt{Equal}(A,B) \\
(\Delta) & P(A)  \leftarrow  Q(A,A) & P(A)  \leftarrow  Q(A,B) \land  \mathtt{Equal}(A,B)\\
\end{array}
$$
Since the equivalent forms above are duplicate-free and two-connected, by (\ref{eq:u}), $ (\nabla),(\Delta)$ can be derived from $\PR_0$, which implies $\mathtt{Claim_2}$.
    \item  Removing the assumption of two-connectedness from $(\ref{eq:k})$, we claim the following inclusion holds:
    \begin{equation}\label{eq:d}
    \mathtt{Claim_3 }: \quad   \mathtt{True},\mathtt{Equal}  \in \Predicates_0 \implies \PR_0[\Predicates] \supset \mathcal{D}_{\Predicates, \leq 2}^{\lbrace 1,2 \rbrace}
     \end{equation}
The additional meta-rules we have to reduce can be narrowed down to:
$$P(A)  \leftarrow  Q(A, A) \land R(A,C) \quad ;  \quad P(A,B)  \leftarrow  Q(A,A) \land R(A,B) \quad ;  \quad  P(A,A)  \leftarrow  Q(A,A) \land R(A,B) 
$$
By using the reduction for duplicated forms $P(A,A), Q(A, A)$ stated previously in $(b)$, these meta-rules may be derived from $ \PR_0$, and $\mathtt{Claim_3}$ follows.\\
    \item In the next step, we extend (\ref{eq:d}) to the connected Horn fragment;
     \begin{equation}\label{eq:c}
      \mathtt{Claim_4}:\quad  \mathtt{True},\mathtt{Equal}  \in \Predicates_0 \implies\PR_0[\Predicates] \supset \mathcal{C}_{\Predicates, \leq 2}^{\lbrace 1,2 \rbrace}
     \end{equation}
    To get rid of the Datalog constraint that each head variable appears in the body, we are brought to examine the following meta-rules:
$$  P(A,B)  \leftarrow  Q(A, C) $$
%P(A)  \leftarrow  Q(B) \quad ;  \quad 
%\quad ;  \quad P(A,B)  \leftarrow  Q(C) 
This meta-rule may be seen as a subcase of $\mathfrak{B}$ once we have matched its second body with $\texttt{True}$.
It ensues that $\PR_0$ is able to generate the fragment $\mathcal{C}_{2}^{\lbrace 1,2 \rbrace}$, as stated in $\mathtt{Claim_4}$.

 \item Lastly, we can get rid of the assumption of "connected".
     \begin{equation}\label{eq:h}
         \mathtt{Claim_5}: \quad  \mathtt{True},\mathtt{Equal}  \in \Predicates_0 \implies \PR_0[\Predicates] \supset \mathcal{F}_{\Predicates, \leq 2}^{\lbrace 1,2 \rbrace} 
     \end{equation}
Upon symmetries and permutations, we are brought to examine the following non-connected meta-rules:
$$
\begin{array}{clll}
 &\text{ Meta-rule }& \text{Reduction} & \text{Comment}\\
(i) & P(A)  \leftarrow  Q(A) \land R(B)  &  \mathfrak{A} & \\
(ii) & P(A)  \leftarrow  Q(A,B) \land R(C)  &  \mathfrak{A} 
 \prescript{}{1}{\odot}\exists &\mathfrak{A}(\exists (Q),R) \\
(iii) & P(A)  \leftarrow  Q(A,B) \land R(C,D)  &\left( \mathfrak{A} 
   \prescript{}{1}{\odot}\exists \right)  \prescript{}{2}{\odot}\exists &\mathfrak{A}(\exists( Q) \land \exists (R)) \\
(iv) & P(A,B)  \leftarrow  Q(A,B) \land R(C)  & \mathfrak{C} 
 \prescript{}{2}{\odot}\rho &\mathfrak{C}(Q,  \rho(R))  \\
(v) & P(A,B)  \leftarrow  Q(A) \land R(B,C)  &   \mathfrak{B} 
 \prescript{}{2}{\odot}\sigma & \mathfrak{B}(Q, \sigma(R)) \\
(vi) & P(A,B)  \leftarrow  Q(A,B) \land R(C,D)  &   \mathfrak{C} 
 \prescript{}{2}{\odot}(\rho\circ \exists ) &\mathfrak{C}(Q,  \rho \circ \exists ( R)))\\
(vii) & P(A,B)  \leftarrow  Q(A,C) \land R(B,D)  &\left(\mathfrak{C} 
 \prescript{}{1}{\odot}\exists \right) \prescript{}{2}{\odot}\sigma &\mathfrak{C}(\exists(Q) \land \sigma (R)))  \\
\end{array}
$$
Note that the clause/operation $\rho$, defined in (\ref{eq:rho},\ref{eq:rho2}),  enables to express the clause $R(C)$ (resp. $R(C,D)$) appearing in the body of $(iv)$ (resp. $(vi)$) as $(\rho)(R)(B)$.
Since, by definition of $(\rho)$, it can be derived from $\PR_0$, we can thereupon conclude that all the above meta-rules are reducible to $\PR_0$, which concludes the proof.
\end{enumerate}
We have therefore proven the following inclusions:
$$\mathtt{True},\mathtt{Equal}  \in \Predicates_0 \implies \PR_0[\Predicates]  = \mathcal{F}_{\Predicates, \leq 2}^{\lbrace 1,2 \rbrace} 	\supset  \mathcal{C}_{\Predicates, \leq 2}^{\lbrace 1,2 \rbrace} \supset \mathcal{D}_{\Predicates, \leq 2}^{\lbrace 1,2 \rbrace} \supset \mathcal{K}_{2}^{\Predicates\lbrace 1,2 \rbrace} 	\supset \mathcal{U}_{\Predicates, \leq 2}^{\lbrace 1,2 \rbrace} $$

\end{proof}

\subsection{ILP Experiment Results}\label{appendix:ILP}

\subsubsection{Extracted Interpretable Solutions}\label{appendix:ILP tasks description}
~\\

We give a detailed description of the ILP tasks in our experiments, \xf{including the target, background knowledge, positive/negative examples, and the learned solution by our method for each task. 
Note that \pw{the} solution for \textit{Fizz} is missing since \pw{our current approach does not solve it with the generic proto-rules in $\PR_*$.}}

\par\noindent \textbf{Predecessor} \quad
The goal of this task is to learn the $predecessor(X,Y)$ relation from examples. 
The background knowledge is the set of facts defining predicate \textit{zero} and the successor relation \textit{succ} 
\[
\mathcal{B}=\{\mathtt{True},\mathtt{False}, zero(0), succ(0, 1), succ(1, 2), \dots\}.
\]

The set of positive examples is 
\[
\mathcal{P}=\{target(1, 0), target(2, 1), target(3, 2), \dots\}
\]
and the set of negative examples is 
\[\mathcal{N}=\{target(X,Y)|(X,Y)\in \{ 0,1,\dots \}^2 \}-\mathcal{P}.
\]
Among these examples, \textit{target} is the name of the target predicate to be learned.
For example, in this task, \textit{target} = \textit{predecessor}.
We use fixed training data for this task given the range of integers.
One solution found by our method is:
\begin{equation*}
    target(X,Y) \leftarrow succ(X,Y).
\end{equation*}

\par\noindent \textbf{Undirected Edge} \quad
A graph is represented by a set of \textit{edge(X,Y)} atoms which define the existence of an edge from node $X$ to $Y$.
The goal of this task is to learn the $undirected\text{-}edge(X,Y)$ relation from examples. 
This relation defines the existence of an edge between nodes $X$ and $Y$ regardless of the direction of the edge.
An example set of background knowledge is 
\[
\mathcal{B}=\{True, False, edge(a,b), edge(b,c)\}.
\] 
The corresponding set of positive examples is 
\[
\mathcal{P}=\{target(a, b), target(b, a), target(b,c), target(c,b)\}
\]
and the set of negative examples is
\[
\mathcal{P}=\{target(a, c), target(c, a)\}.
\]

We use randomly generated training data for this task given the number of nodes in the graph.
One solution found by our method is:
\begin{align*}
target(X,Y) & \leftarrow \left( aux1(X,Y) \land edge(Y,X) \right) \vee edge(X,Y), \\
aux1(X,Y) & \leftarrow edge(X,Y),
\end{align*}
where $aux1$ is an invented auxiliary predicate.

\par\noindent \textbf{Less Than} \quad
The goal of this task is to learn the $less\text{-}than(X,Y)$ relation which is true if $X$ is less than $Y$.
Here the background knowledge is the same as that in Predecessor task. 
The set of positive examples is 
\[
\mathcal{P}=\{target(X,Y)|X<Y \}
\]
and the set of negative examples is 
\[
\mathcal{N}=\{target(X,Y)| X \ge Y \}.
\]

We use fixed training data for this task given the range of integers.
One solution found by our method is:
\begin{align*}
    target(X,Y) & \leftarrow \left( target(X,Z) \land target(Z,Y) \right) \vee succ(X,Y) .
\end{align*}

\par\noindent \textbf{Member} \quad
The goal of this task is to learn the $member(X,Y)$ relation which is true if $X$ is an element in list $Y$.
Elements in a list are encoded with two relations $cons$ and $values$, where $cons(X,Y)$ if $Y$ is a node after $X$ with null node $0$ being the termination of lists and $value(X,Y)$ if $Y$ is the value of node $X$.

Take the list $[3,2,1]$ as an example.
% The background knowledge is $\mathcal{B}=\{ \}$. 
The corresponding set of positive examples is 
\begin{equation*}
    \begin{split}
        \mathcal{P}=\{ & target(3,[3,2,1]),\quad target(2,[3,2,1]), \quad target(1,[3,2,1]),\\
        & target(2,[2,1]),\quad target(1,[2,1],\quad target(3,[3,2]), \\
        & target(2,[3,2]),\quad target(3,[3,1]),\quad target(1, [3,1]), \\
        & True, False\}
    \end{split}
\end{equation*}
and the set of negative examples is
\[
\mathcal{P}=\{target(3, [2,1]), target(1, [3,2]), target(2,[3,1])\}.
\]

We use randomly generated training data for this task given the length of the list.
One solution found by our method is:
\begin{align*}
    % target(X,Y) & \leftarrow value(Y,X), \\
    % target(X,Y) & \leftarrow cons(Y,Z) \land target(X,Z).
    target(X,Y) & \leftarrow \left( target(X,Z) \land cons(Z,Y) \right) \vee value(X,Y).
\end{align*}

\par\noindent \textbf{Connectedness} \quad
The goal of this task is to learn the $connected(X,Y)$ relation which is true if there is a sequence of edges connecting nodes $X$ and $Y$.
An example set of background knowledge is
\[
\mathcal{B}=\{True, False, edge(a,b), edge(b,c), edge(c,d) \}.
\]
The corresponding set of positive examples is 
\[
\mathcal{P}=\{target(a,b), target(b,c), target(c,d), target(a,c), target(a,d), target(b,d)\}
\]
and the set of negative examples is 
\[
\mathcal{P}=\{target(b, a), target(c, b), target(d,c), target(d,a), target(d,b), target(c,a)\}.
\]

We use randomly generated training data for this task given the number of nodes in the graph.
One solution found by our method is:
\begin{align*}
% target(X,Y) & \leftarrow edge(X,Y), \\
% target(X,Y) & \leftarrow edge(X,Z) \land target(Z,Y).
target(X,Y) & \leftarrow \left( target(X,Z) \land target(Z,Y) \right) \vee edge(X,Y).
\end{align*}

\par\noindent \textbf{Son} \quad
The goal of this task is to learn the $son\text{-}of(X,Y)$ relation which is true if $X$ is the son of $Y$.
The background knowledge consists of various facts about a family tree containing the relations $father\text{-}of, bother\text{-}of$ and $sister\text{-}of$. 
An example set of background knowledge is
\[
\mathcal{B}=\{True, False, father(a,b),father(a,c),father(a,d),bother(b,c),bother(d,c),sister(c,b) \}.
\]
The corresponding set of positive examples is 
\[
\mathcal{P}=\{ target(b,a), target(d,a) \}
\]
and the set of negative examples $\mathcal{N}$ is a subset of all ground atoms involving the target predicates that are not in $\mathcal{P}$.

We use randomly generated training data for this task given the number of nodes in the family tree.
One solution found by our method is:
\begin{align*}
    % target(X,Y) & \leftarrow father(Y,X) \land aux1(X) \\
    % aux1(X) & \leftarrow brother(X,Y) \\
    % aux1(X) & \leftarrow father(X,Y), 8
    target(X,Y) & \leftarrow \left( aux1(X,Y) \land father(Y,X) \right) \vee False \\
    aux1(X) & \leftarrow \left( father(X,Z) \land True \right) \vee brother(X,T)
\end{align*}
where $aux1$ is an invented auxiliary predicate.

\par\noindent \textbf{Grandparent} \quad
The goal of this task is to learn the $grandparent(X,Y)$ relation which is true if $X$ is the grandparent of $Y$.
The background knowledge consists of various facts about a family tree containing the relations $father\text{-}of \text{ and } mother\text{-}of$.
An example set of background knowledge is
\[
\mathcal{B}=\{ father(c,b),father(b,a),mother(d,b),mother(e,a), True, False \}.
\]
The corresponding set of positive examples is 
\[
\mathcal{P}=\{ target(c,a), target(d,a) \}
\]
and the set of negative examples $\mathcal{N}$ is a subset of all ground atoms involving the target predicates that are not in $\mathcal{P}$.

We use randomly generated training data for this task given the number of nodes in the family tree.
One solution found by our method is:
\begin{align*}
    target(X,Y) & \leftarrow \left( aux1(X,Z) \land aux1(Z,Y) \right) \vee False, \\
    aux1(X,Y) & \leftarrow \left( mother(X,Y) \land True \right) \vee father(X,Y), 
\end{align*}
where $aux1$ is an invented auxiliary predicate.

\par\noindent \textbf{Adjacent to Red} \quad
In this task, nodes of the graph are colored either green or red.
The goal of this task is to learn the $is\text{-}adjacent\text{-}to\text{-}a\text{-}red\text{-}node(X)$ relation which is true if node $X$ is adjacent to a red node.
Other than the relation $edge$, the background knowledge also consists of facts of relations $colour \text{ and } red$, where $colour(X,C)$ if node $X$ has colour $C$ and $red(C)$ if colour of $C$ is red.

An example set of background knowledge is 
\[
\mathcal{B}=\{True, False, edge(a,b),edge(b,a),edge(d,e),edge(d,f),colour(a,p), red(p), colour(d,q), red(q) \}.
\]
The corresponding set of positive examples is
\[
\mathcal{P}=\{ target(b), target(e), target(f) \}
\]
and the set of negative examples $\mathcal{N}$ is a subset of all ground atoms involving the target predicates that are not in $\mathcal{P}$.

We use randomly generated training data for this task given the number of nodes in the graph.
One solution found by our method is:
\begin{align*}
    % target(X) & \leftarrow edge(X,Y) \land aux1(Y), \\
    % aux1(X) & \leftarrow colour(X,Y) \land red(Y), 
    % red edge colour
    target(X) & \leftarrow \left( edge(X,Z) \land aux1(Z,X) \right) \vee False, \\
    aux1(X) & \leftarrow \left( colour(X,Z) \land red(Z,X) \right) \vee False,
\end{align*}
where $aux1$ is an invented auxiliary predicate.

\par\noindent \textbf{Two Children} \quad
The goal of this task is to learn the $has\text{-}at\text{-}least\text{-}two\text{-}children(X)$ relation which is true if node $X$ has at least two child nodes.
Other than the relation $edge$, the background knowledge also consists of facts of \textit{not-equals} relation $neq$, where $neq(X,Y)$ if node $X$ does not equal to node $Y$.

An example set of background knowledge is 
\[
\mathcal{B}=\{True, False, edge(a,b),edge(a,c),edge(c,d),neq(a,b),neq(a,c),neq(a,d),neq(b,c),neq(b,d),neq(c,d) \}.
\]
The corresponding set of positive example(s) is 
\[
\mathcal{P}=\{ target(a) \}
\]
and the set of negative examples $\mathcal{N}$ is a subset of all ground atoms involving the target predicates that are not in $\mathcal{P}$.

We use randomly generated training data for this task given the number of nodes in the graph.
One solution found by our method is:
\begin{align*}
    % target(X) & \leftarrow edge(X,Y) \land aux1(X,Y), \\
    % aux1(X,Y) & \leftarrow edge(X,Z), neq(Z,Y),
    % neq, edge
    target(X) & \leftarrow \left( aux1(X,Z) \land edge(Z,X) \right) \vee False , \\
    aux1(X,Y) & \leftarrow \left( edge(X,Z), \land neq(Z,Y)\right) \vee False ,
\end{align*}
where $aux1$ is an invented auxiliary predicate.

\zj{\par\noindent \textbf{Relatedness} \quad
The goal of this task is to learn the $related(X, Y)$ relation, which is true if two node\xf{s} $X$ and $Y$ \xf{have} \pw{any} family relations. 
The background knowledge \xf{is} $parent(X, Y)$ if $X$ is $Y$'s parent.

An example set of background knowledge is
\[
\mathcal{B}=\{True, False, parent(a, b), parent(a, c), parent(c, e), parent(c, f), parent(d, c), parent(g, h)\}.
\]
The corresponding set of positive examples is 
\begin{equation*}
    \begin{split}
        \mathcal{P}=\{ &target(a, b), target(a, c), target(a, d), target(a, e), target(a, f),\\
            & target(b, a), target(b, c), target(b, d), target(b, e), target(b, f),\\
            & target(c, a), target(c, b), target(c, d), target(c, e), target(c, f),\\
            & target(d, a), target(d, b), target(d, c), target(d, e), target(d, f),\\
            & target(e, a), target(e, b), target(e, c), target(e, d), target(e, f),\\
            & target(f, a), target(f, b), target(f, c), target(f, d), target(f, e)\\
            & target(g, h), target(h, g)\}
    \end{split}
\end{equation*}
and the set of negative examples $\mathcal{N}$ is a subset of all ground atoms involving the target predicates that are not in $\mathcal{P}$.

We use randomly generated training data for this task given the number of nodes in the \xf{family tree}.
One solution found by our method is:
\begin{align*}
    % target(X) & \leftarrow edge(X,Y) \land aux1(Y), \\
    % aux1(X) & \leftarrow colour(X,Y) \land red(Y), 
    % red edge colour
     target(X,Y) & \leftarrow \left( target(X,Z) \land aux1(Z,Y) \right) \vee parent(X, Y)\xf{,} \\
     aux1(X,Y) & \leftarrow \left( target(X,Y) \land target(Y,X) \right) \vee parent(X, Y)\xf{,} 
\end{align*}
where $aux1$ is an invented auxiliary predicate.
}
\par\noindent \textbf{Cyclic} \quad
The goal of this task is to learn the $is\text{-}cyclic(X)$ relation which is true if there is a path, i.e., a sequence of \textit{edge} connections, from node $X$ back to itself.
An example set of background knowledge is 
\[
\mathcal{B}=\{True, False, edge(a,b),edge(b,a),edge(d,c),edge(d,b) \}.
\]
The corresponding set of positive examples is 
\[
\mathcal{P}=\{ target(a),target(b) \}
\]
and the set of negative examples $\mathcal{N}$ is a subset of all ground atoms involving the target predicates that are not in $\mathcal{P}$.

We use randomly generated training data for this task given the number of nodes in the graph.
One solution found by our method is:
\begin{align*}
    % target(X) & \leftarrow aux1(X,X), \\
    % aux1(X,Y) & \leftarrow edge(X,Y), \\
    % aux1(X,Y) & \leftarrow aux1(X,Z) \land aux1(Z,Y),
    target(X) & \leftarrow \left( aux1(X,Z) \land aux1(Z,X) \right) \vee False, \\
    aux1(X,Y) & \leftarrow \left( aux1(X,Z) \land edge(Z,Y) \right) \vee edge(X,Y).
\end{align*}
where $aux1$ is an invented auxiliary predicate.

\par\noindent \textbf{Graph Coloring} \quad
The goal of this task is to learn the $adj\text{-}to\text{-}same(X,Y)$ relation which is true if nodes $X$ and $Y$ are of the same colour and there is an edge connection between them.
The background knowledge consists of facts about a coloured graph containing the relations $edge \text{ and } colour$, which are similar to those in the task \textit{Adjacent to Red}.
An example set of background knowledge is 
\[
\mathcal{B}=\{True,False, edge(a,b),edge(b,a),edge(b,c),edge(a,d),colour(a,p),colour(b,p),colour(c,q),colour(d,q) \}.
\]
The corresponding set of positive examples is 
\[
\mathcal{P}=\{ target(a,b), target(b,a) \}
\]
and the set of negative examples $\mathcal{N}$ is a subset of all ground atoms involving the target predicates that are not in $\mathcal{P}$.

We use randomly generated training data for this task given the number of nodes in the graph.
One solution found by our method is:
\begin{align*}
    % target(X,Y) & \leftarrow edge(X,Y) \land aux1(X,Y), \\
    % aux1(X,Y) & \leftarrow colour(X,Z) \land colour(Y,Z),
    % edge, colour
    target(X,Y) & \leftarrow \left( edge(X,Z) \land aux1(Z,X) \right) \vee False, \\
    aux1(X,Y) & \leftarrow \left( colour(X,Z) \land colour(Z,Y) \right) \vee False,
\end{align*}
where $aux1$ is an invented auxiliary predicate.

\par\noindent \textbf{Length} \quad
The goal of this task is to learn the $length(X, Y)$ relation which is true if the length of list $X$ is $Y$.
Similar to the task \textit{Member}, elements in a list are encoded with two relations $cons$ and $succ$, where $cons(X,Y)$ if $Y$ is a node after $X$ with null node $0$ being the termination of lists and $succ(X,Y)$ if $Y$ is the next value of integer $X$. 
Moreover, this task adds $zero(X)$ as another background predicate, which is true if $X$ is $0$.

Take the list $[3,2,1]$ as an example, suppose node $0$ is the end of a list.
The background knowledge is 
\[
\mathcal{B}=\{ True, False, zero(0), succ(0, 1), succ(1, 2), succ(2, 3)\}. 
\]
The corresponding set of positive examples is 
\begin{equation*}
    \begin{split}
        \mathcal{P}=\{ & target([3,2,1], 3),\quad target([2,1], 2), \quad target([1], 1)\}
    \end{split}
\end{equation*}
We use randomly generated training data for this task given the number of nodes in a list.

\par\noindent \textbf{Even-Odd} \quad
The goal of this task is to learn the $even(X)$ relation which is true if value $X$ is an even number.
The background knowledge includes two predicates, one is $zero(X)$, which is true if $X$ is $0$, another one is $succ(X, Y)$, which is true if $Y$ is the next value of $X$.
An example set of background knowledge is 
\[
\mathcal{B}=\{True, False, zero(0), succ(0, 1), succ(1, 2), \dots\}.
\]
The corresponding set of positive examples is 
\[
\mathcal{P}=\{ target(0), target(2), target(4), \dots \}
\]
and the set of negative examples is 
\[
\mathcal{N}=\{target(1), target(3), target(5), \dots \}.
\]
Once the number of constants is given, the dataset is deterministic.
One solution found by our method is:
\xf{
\begin{align*}
  target(X) & \leftarrow \left( zero(X) \land aux1(Z,X) \right) \vee zero(X), \\
  aux1(X,Y) & \leftarrow \left( aux1(X,Z) \land aux1(Z,Y) \right) \vee aux2(X,Y), \\
  aux2(X,Y) & \leftarrow \left( succ(X,Z) \land succ(Z,Y) \right) \vee False,
\end{align*}
where $aux1$ and $aux2$ are invented auxiliary predicates.}

\noindent \textbf{Even-Succ2} \quad
\zj{Even-succ has \pw{the} same backgrounds and target predicates \pw{as} Even-Odd. 
In \citet{campero2018logical}, they provide two different templates set \cg{tailored to} Even-Succ  \cg{respectively to} Even-Odd. \cg{However, in our approach}, this difference \cg{is not relevant anymore}: since we provide a uniform template set, these two tasks become identical.}

\noindent \textbf{Buzz} \quad
The goal of this task is to learn the $buzz(X)$ relation which is true if value $X$ is divisible by 5.
The background knowledge consists of 4 predicates:
$zero(X)$ is true if $X$ is $0$, $succ(X, Y)$ is true if $Y$ is the next value of $X$, $pred1(X, Y)$ is true if $Y=X+3$, $pred2(X, Y)$ is true if $Y=X+2$. An example set of background knowledge is
\[
\mathcal{B}=\{True, False, zero(0), succ(0, 1), succ(1, 2), \dots, pred1(0, 3), pred2(2, 4), \dots, pred2(0, 2), pred2(1, 3), \dots\}.
\]
The corresponding set of positive examples is 
\[
\mathcal{P}=\{ target(0), target(5), \dots \}
\]
and the set of negative examples is 
\[
\mathcal{N}=\{target(1), target(2), target(3), target(4), target(6), target(7), \dots \}.
\]
Once the number of constants is given, the dataset is deterministic. 
One solution found by our method is:
\xf{
\begin{align*}
  target(X) & \leftarrow \left( aux1(X,Z) \land pred2(Z,X) \right) \vee zero(X), \\
  aux1(X,Y) & \leftarrow \left( aux2(X,Z) \land pred1(Z,Y) \right) \vee False, \\
  aux2(X,Y) & \leftarrow \left( True \land aux3(Y) \right) \vee zero(X), \\
  aux3(X) & \leftarrow \left( aux1(X,Z) \land pred2(Z,X) \right) \vee zero(X),
\end{align*}
where $aux1,$ $aux2$ and $aux3$ are invented auxiliary predicates.}

\par\noindent \textbf{Fizz} \quad
The goal of this task is to learn the $fizz(X)$ relation which is true if value $X$ is divisible by 3.
The background knowledge is the same with $Even-Odd$ task.
The corresponding set of positive examples is 
\[
\mathcal{P}=\{ target(0), target(5), \dots \}
\]
and the set of negative examples is 
\[
\mathcal{N}=\{target(1), target(2), target(3), target(4), target(6), target(7), \dots \}.
\]
Once the number of constants is given, the dataset is deterministic.

\subsubsection{Further Experimental Results}\label{appendix:ILP_further_results}
~\\

In Table~\ref{table:acc-result all}, we provide the results for all the ILP tasks.

\begin{table}[H]
\centering
\begin{tabular}{cccccccc}
\toprule
\multirow{2}{*}{Task} & \multirow{2}{*}{$|I|$} & \multirow{2}{*}{Recursive} & \multirow{2}{*}{$\partial$ILP} & \multirow{2}{*}{LRI} & \multicolumn{3}{c}{Ours}                        \\
\cmidrule{6-8}
                      &                      &                            &                           &                           & train & soft evaluation & symbolic evaluation \\
                      \midrule
Predecessor           & 1                    & No                         & 100                       & 100                       & 100   & 100               & 100                 \\
Undirected Edge       & 1                    & No                         & 100                       & 100                       & 100   & 100               & 100                 \\
Less than             & 1                    & Yes                        & 100                       & 100                       & 100   & 100               & 100                 \\
Member                & 1                    & Yes                        & 100                       & 100                       & 100   & 100               & 100                 \\
Connectedness         & 1                    & Yes                        & 100                       & 100                       & 100   & 100               & 100                 \\
Son                   & 2                    & No                         & 100                       & 100                       & 100   & 100               & 100                 \\
Grandparent           & 2                    & No                         & 96.5                      & 100                       & 100   & 100               & 100                 \\
Adjacent to Red       & 2                    & No                         & 50.5                      & 100                       & 100   & 100               & 100                 \\
Two Children          & 2                    & No                         & 95                        & 0                         & 100   & 100               & 100                 \\
Relatedness           & 2                    & Yes                        & 100                       & 100                       & 100   & 100               & 100                 \\
Cyclic                & 2                    & Yes                        & 100                       & 100                       & 100   & 100               & 100                 \\
Graph Coloring        & 2                    & Yes                        & 94.5                      & 0                         & 100   & 100               & 100                 \\
Length                & 2                    & Yes                        & 92.5                      & 100                       & 20    & 0                 & 0                   \\
Even-Odd              & 2                    & Yes                        & 100                       & 100                       & 40    & 40                & 40                   \\
Even-Succ2            & 2                    & Yes                        & 48.5                      & 100                       & 40    & 40                & 40                  \\
Buzz                  & 2                    & Yes                        & 35                        & 70                        & 100   & 40                & 40                  \\
Fizz                  & 3                    & Yes                        & 10                        & 10                        & 0     & 0                 & 0  \\
\bottomrule
\end{tabular}
\caption{Percentage of successful runs among 10 runs. \xf{$|I|$} is the smallest number of intensional predicates needed. Recursive means whether or not the solution need\pw{s} to learn recursive rules.}
\label{table:acc-result all}
\end{table}

About \texttt{Length} Task:  
The deceiving performance in this table for task \texttt{Length} can be easily explained: in theory our model corresponds to rule-induction with function-free definite Horn clause in theory; yet, de facto, in the recursive-case, both the number of layers and the number of instantiated auxiliary predicate by proto-rule per layer define the actual expressivity of the model.
The number of of instantiated auxiliary predicate by proto-rule per layer is by default set at $1$ in all our experiments, as we intended to share the same hyperparameters set on all tasks;
however, to have a chance to solve the task \texttt{Length}, we should increase this number to $2$, to widen the hypothesis space. 

About \texttt{Even-Odd} Task:  
\cg{As mentionned above,  task \textit{Even-Odd} corresponds, for our method, to the same task \pw{as} the task \textit{Even-Succ2}, as the target predicate is identical. 
This is not the case for other ILP approaches like $\partial$ILP or LRI, as they hand-engineer different template set\pw{s} for each of these two tasks, corresponding to the desired auxiliary predicate. 
}%
% For the ILP tasks that our model could solve (i.e., all except \textit{Length} and \textit{Fizz}), we also provide the training times averaged over 10 runs in Table~\ref{table:time-result}.

% \begin{table}[H]
% \centering
% \begin{tabular}{cccc}
% \toprule
% Task & $|I|$ & Recursive & Ours\\
%                       \midrule
% Predecessor           & 1  & No       & 893    \\  % 893.07
% Undirected Edge       & 1  & No       & 104    \\  % 104.80
% Less than             & 1  & Yes      & 2680   \\  % 2680.01
% Member                & 1  & Yes      & 278    \\  % 278.33
% Connectedness         & 1  & Yes      & 268     \\  % 268.50
% Son                   & 2  & No       & 963    \\  % 963.78
% Grandparent           & 2  & No       & 1227   \\  % 1227.13
% Adjacent to Red       & 2  & No       & 494    \\  % 494.81
% Two Children          & 2  & No       & 1944   \\  % 1944.97
% Relatedness           & 2  & Yes      & 3959   \\  % 3959.53
% Cyclic                & 2  & Yes      & 483    \\  % 483.97
% Graph Coloring        & 2  & Yes      & 5592   \\  % 5592.09
% %Length                & 2  & Yes      & --\\
% Even-Odd              & 2  & Yes      & 6517   \\  % 6517.95
% Even-Succ2            & 2  & Yes      & 10238  \\  % 10238
% Buzz                  & 2  & Yes      & 28932  \\  % 28932.41
% %Fizz                  & 3  & Yes      & --\\
% \bottomrule
% \end{tabular}
% \caption{Average training time(s) of 10 runs. 
% % \zj{We didn't put the results for $Length$ and $Fizz$ for now, since we haven't get a good performance for them.} 
% \label{table:time-result}
% \end{table}

\subsubsection{Operation Choices}\label{appendix:operation choices}
~\\

 \zj{In our implementation, the default choice is sum for $\textsc{POOL}$, $\min$ for $\textsc{AND}$, $\max$ for $\textsc{OR}$. \cg{In \ref{table:res_ILP_POOL_MERGE_OR_AND}, we experimentally compare different choices for these operations.} The first column shows the results with our default choices, while other columns show the results by using max for $\textsc{POOL}$, product for $\textsc{AND}$, and prodminus\footnote{$prodminus(v_1, v_2) = v_1+v_2-v_1*v_2$} for $\textsc{OR}$.}
 
\begin{table}[H]
\centering
\begin{tabular}{cccccc}
\toprule
    Task      & default & POOL-max & AND-product & OR-prodminus \\
\midrule
Adjacent To Red & 100 & 20 & 100 & 100 \\
Member          & 100 & 40 & 100 & 90  \\
Cyclic          & 100 & 50 & 90  & 100 \\
Two Children    & 100 & 70 & 80  & 80  \\
\bottomrule
\end{tabular}
\caption{\zj{Percentage of successful runs among 10 runs using soft evaluation, obtained by models trained with different implementations for $\textsc{POOL}$, $\textsc{MERGE}$, $\textsc{AND}$, $\textsc{OR}$.}}
\label{table:res_ILP_POOL_MERGE_OR_AND}
\end{table}

\subsubsection{Limitations of LRI \citep{campero2018logical}}\label{sec:limitations LRI}
~\\

% While LRI can obtain good results as indicated in Table~\ref{table:acc-result all} if provided with meta-rules customized for each task, the performance quickly degrades when provided a set of more generic meta-rules (Table~\ref{table:LRI_rule_union_res}).

\zj{Using LRI, \cg{if gathering all the templates needed for all ILP tasks (in Table \ref{table:acc-result all}) in a template set \cg{$\PR_{LRI}$}, we obtain $18$ templates}. Table \ref{table:LRI_rule_union_res} demonstrates how the evaluation success rate decrease if trained LRI with $R_{LRI}$. While LRI can obtain good results as indicated in Table~\ref{table:acc-result all} if provided with meta-rules customized for each task, the performance quickly degrades for hard tasks when provided a set of more generic meta-rules. \cg{This phenomenon is more accentuated for more complex tasks.} For easy tasks, like Predecessor, the performance didn't change.
% : we tested it by randomly sampling rule templates from the union set \cg{$\PR_{LRI}$}, while making sure the sampled template set include the specific necessary templates for the task, as provided in \citep{campero2018logical}. In order to have a fair competition, we also use randomly generated data to train LRI for stochastic tasks.
% It's noticed that for easy tasks like Predecessor, Cyclic and Member, LRI still can get a relatively good performance, while for  hard tasks like EvenSucc and Adjacent To Red the decrease is obvious.
}
\begin{table}[H]
\centering
\begin{tabular}{ccc}
\toprule
Tasks & LRI with specific templates & LRI with $R_{LRI}$ \\
\midrule
Predecessor           & 100 & 100 \\
Undirected Edge       & 100 & 80 \\
Less than             & 100 & 100 \\
Member                & 100 & 30 \\
Connectedness         & 100 & 100 \\
Son                   & 100 & 80 \\
Grandparent           & 100 & 90 \\
Adjacent to Red       & 100 & 0 \\
Two Children          &  0 &  0\\
Relatedness           &  100 & 100\\
Cyclic                & 100 & 0 \\
Graph Coloring        & 0 &  0 \\
Length                & 100 & 50\\
Even-Odd              & 100 & 20 \\
Even-Succ2            & 100 & 10 \\
Buzz                  & 70 & 0\\
Fizz                  & 10 & 0\\
\bottomrule
\end{tabular}
\caption{ LRI's performance with increasing number of rule templates. Mearsured by percentage of successful runs among 10 runs using soft evaluation.}
\label{table:LRI_rule_union_res}
\end{table}
% \begin{table}[H]
% \centering
% \begin{tabular}{ccccc}
% \toprule
% task & LRI & TODO
% \multirow{2}{*}{task} & \multicolumn{4}{c}{\#rule-templates} \\
% \cmidrule{2-5}
%                       & 5      & 10      & 15      & 18      \\
%                       \midrule
%  Predecessor          & 100    & 100     & 100    & 100     \\
%   Cyclic               & 100    & 90      & 90     & 70      \\
%   Member              & 100    & 90      & 80     & 70      \\
%  EvenSucc             & 80     & 80      & 80     & 40      \\
%  AdjToRed             & 80     & 80      & 40     & 0      \\
%  %Buzz                 &      &       &      &       \\
%  %Member                 &      &       &      &       \\
% \bottomrule
% \end{tabular}
% \caption{ LRI's performance with increasing number of rule templates. Mearsured by percentage of successful runs among 10 runs using soft evaluation.}
% \label{table:LRI_rule_union_res}
% \end{table}

\subsection{Other Experimental Results}\label{appendix:other}
\subsubsection{Visual Genome Experiments}\label{appendix:visual_genome_experiments}
~\\

For this experiment, we use a dataset called GQA \citep{Hudson_Manning_2019} \cg{which is} a preprocessed version of the Visual Genome dataset \citep{krishna_visual_2017}, \cg{since the original is commonly} considered to be too noisy \citep{Zellers_Yatskar_Thomson_Choi_2018}.
In GQA, the original scene graphs have been converted into a collection of KBs, leading to $1.9$M facts, $1.4$M constants, and $2100$ predicates.
Following \citep{yang_learn_2019}, we filter this dataset to remove the predicates that appear less than $1500$ times and to focus on the top $150$ objects.

We train $150$ models to provide a logic explanation to those $150$ objects.
For the evaluation metrics, we use recall $@1 (R@1)$ and recall $@5 (R@5)$, which are computed on a held-out set.
$R@k$ measures the fraction of ground-truth atoms that appear among the top $k$ most confident predictions in an image.
In our model, even though all the models are instantiated with the same max layer $n_L$, the trained model may have different layers. 
Indeed, since each auxiliary predicate at layer $\ell$ may be formed with predicates from any layer $0$ to $\ell$ in  $\Predicates^\downarrow_\ell$), the trained model may contain $1$ to $n_L$ layers (without counting the target predicate).
Given the soft unification computation in \eqref{equ:one_step_inference}, a trained model with a larger number of layers has a tendency to output smaller values.
Therefore, to make the output of all the models comparable, we use a simple $L_2$ normalization before comparing the outputs of those trained models in order to compute the evaluation metrics.

\zj{In the multi-task setting we initialized background embeddings from GPT2. \cg{We compare the semantic space underlying these embeddings after fine-tuning (i.e. training of our model).}  More precisely, we use cosine similarity to measure distances between all pairs of background embeddings, sort and select top 10 closest pairs. As Table \ref{table:close_embeddings_MTGQA} \cg{suggests}, \cg{the fine-tuned embeddings pairs have akin similarities;}  this initialization choice may therefore help with the performance \cg{to} some extent.}
\begin{table}[H]
\centering
\begin{tabular}{cc}
\toprule
    & Top 10 pairs of similar embeddings \\
\midrule
\multirow{2}{*}{GPT2} & (bag, backpack), (arrow, apple), (airplane, air), (backpack, airplane), (apple, airplane),\\
 & (animal, airplane), (arrow, animal), (at, above), (apple, animal), (bag, airplane) \\
 \midrule
\multirow{2}{*}{Fine-tuned} & (bag, backpack), (bag, airplane), (arrow, apple), (airplane, air), (arm, air),\\
& (backpack, airplane), (air, above), (apple, air), (arrow, animal), (arm, airplane)\\
\bottomrule
\end{tabular}
\caption{Top 10 closest embeddings from GPT2 and our trained multi-task model.}
\label{table:close_embeddings_MTGQA}
\end{table}

\subsubsection{RL Experiments}\label{appendix:rl_experiments}
~\\

We also tested our model in a sequential decision-making framework: 
on the tasks Stack, Unstack and On from the Blocksworld environment, as described in \citet{jiang_neural_2019}.
In those tasks, an agent has 50 steps to build a stack (Stack task), place all the blocks directly on the floor (Unstack task), or move a specific block into another (On task).

From all the valuations of the target predicate Move(X,Y), we compute a softmax policy used during the exploration. 
During training, the supervised BCE loss \eqref{eq:bceloss} is replaced by a standard PPO loss \citep{SchulmanWolskiDhariwalRadfordKlimov17} on the softmax policy.
To estimate the advantage function, we relied on the same critic architecture defined by \citet{zimmer2021differentiable}.

At the end of the training, the symbolic policy is extracted and evaluated on 5 testing scenarios not seen during training. We reported the performance in Table~\ref{table:RL-result}. Our model can be better than NLRL and as good as NLM and DLM.
\begin{table*}[h]
\centering
\parbox{.6\linewidth}{
\centering
\begin{tabular}{@{}ccccccccc@{}}
\toprule
\multirow{2}{*}{Task} & 
\multicolumn{4}{c}{Rewards} &           
\multicolumn{4}{c}{Training time}                        \\
\cmidrule{2-9}
                      & NLRL & NLM & DLM & Ours & NLRL & NLM & DLM & Ours \\
\midrule
Unstack & 0.914 & 0.920 & 0.920 & 0.920 & \multirow{3}{*}{hours} & \multirow{3}{*}{minutes} & \multirow{3}{*}{minutes} & \multirow{3}{*}{minutes} \\ 

%\midrule
Stack        & 0.877 & 0.920 & 0.920 & 0.920 \\ 
%\midrule
On        & 0.885 & 0.896 & 0.896 & 0.896  \\
\bottomrule
\end{tabular}
\caption{Comparisons with NLRL/NLM/DLM in terms of rewards on the testing scenarios and the order of magnitude of training times.}
\label{table:RL-result}
}
\end{table*}

\subsubsection{Sensitivity to Hyperparameters}
~\\

\zj{\cg{To evaluate the sensitivity of our model to hyperparameter, we tested} other hyperparameter choices on some ILP tasks.
\cg{We refer to Table} \ref{table:spe-hyper-param} for the results obtained with default hyperparameters,  presented Table \ref{table:generic-hyper-param}.  As Table \ref{table:ILP_hyperparameter_sensitivity_res} suggests, both smaller inference steps, or smaller depth will affect performance; 
\cg{this performance decrease natural, since it narrows down  the hypothesis space, which may not contain anymore the solution needed for the task.
}
\cg{However, we can appreciate the fact that  larger inference step or depth usually will not affect much the performance (until a limit).
}
\cg{We can also notice} that cosine performs better than other similarity functions; 
\cg{naturally, restricting or excluding the} recursivity would perform well compared with full recursivity, unless the Task require it.
% \begin{item}
% \item Usually smaller inference steps will decrease the performance, while larger inference steps may not.
% \item Cosine function works better than other similarity functions.
% \item Add recursivity to rules would be better.
% \item 
% \end{item}
}

\begin{table}[H]
\centering
\begin{tabular}{cccccccccc}
\toprule
\multirow{2}{*}{task} & \multirow{2}{*}{default} & \multicolumn{8}{c}{inference-steps (train-steps, eval-steps)} \\
\cmidrule{3-10}
 & & \multicolumn{2}{c}{$(s_t-2, s_e-2)$} & \multicolumn{2}{c}{$(s_t-1, s_e-1)$} & \multicolumn{2}{c}{$(s_t+1, s_e+1)$} & \multicolumn{2}{c}{$(s_t+2, s_e+2)$} \\
\midrule
 Adjacent to Red & 100 & \multicolumn{2}{c}{90} & \multicolumn{2}{c}{100} & \multicolumn{2}{c}{100} & \multicolumn{2}{c}{100} \\
 Member & 100 & \multicolumn{2}{c}{100} & \multicolumn{2}{c}{100} & \multicolumn{2}{c}{100} & \multicolumn{2}{c}{100} \\
 Cyclic & 100 & \multicolumn{2}{c}{90} & \multicolumn{2}{c}{90} & \multicolumn{2}{c}{90} & \multicolumn{2}{c}{100} \\
 Two Children & 100 & \multicolumn{2}{c}{80} & \multicolumn{2}{c}{100} & \multicolumn{2}{c}{100} & \multicolumn{2}{c}{100} \\
\midrule
\midrule
\multirow{2}{*}{task} &
\multicolumn{3}{c}{similarity}  & \multicolumn{2}{c}{recursivity} & \multicolumn{4}{c}{max-depth} \\
\cmidrule{2-10}
 & L1 & L2 & scalar-product & none & iso-recursive  & 1 & 2 & 3 & 5 \\
\midrule
 Adjacent to Red & 10 & 80 & 40 & 100 & 100 & 0 & 90 & 100 & 100   \\
 Member & 60 & 100 & 80 & 0 & 100 & 50 & 100 & 100 & 100 \\
 Cyclic & 60 & 80 & 50 & 90 & 100 & 10 & 60 & 80 & 90 \\
 Two Children & 0 & 60 & 10 & 90 & 100 & 0 & 70 & 70 & 100  \\
\bottomrule
\end{tabular}
\caption{\zj{Percentage of successful runs among 10 runs using soft evaluation, obtained by models trained with different hyper-parameter choices. Here, $s_t$ and $s_e$ are default inference steps used in training and evaluation, shown in Table \ref{table:spe-hyper-param}.}}
\label{table:ILP_hyperparameter_sensitivity_res}
\end{table}

\subsection{Hyperparameters}\label{appendix:Hyper-param}

\xf{
We list relevant generic and task-specific hyper-parameters used for our training method in \pw{Tables}~\ref{table:generic-hyper-param} and \ref{table:spe-hyper-param}, respectively.
In \pw{Table}~\ref{table:spe-hyper-param}, the hyper-parameters \textit{train-num-constants} and \textit{eval-num-constants} represent the number of constants during training and evaluation, respectively.
We keep the values of the two hyper-parameters for each task \pw{the} same as those used in \cite{campero2018logical}.
\cg{Note that our model do not require the actual knowledge of the depth of the solution; the \textit{max-depth} parameter simply could be an upper bound.
Although, this parameter could be reduced for simpler tasks,  we set \textit{max-depth}$=4$ for all tasks to make our training method more generic.
}
}
\begin{table}[H]
\centering
\begin{tabular}{ccccccc}
\toprule
\textbf{Hyper-parameter} & recursivity & fuzzy-and & fuzzy-or & similarity & lr & lr-rules  \\
\midrule
\textbf{Value}           & full & min & max & cosine & 0.01 & 0.03   \\
\midrule
\midrule
\textbf{Hyper-parameter} & temperature & \pw{G}umbel-noise & \pw{G}umbel-noise-decay-mode &  &  &   \\
\midrule
\textbf{Value}           & 0.1 & 0.3 & linear &  &  &    \\
\bottomrule
\end{tabular}
\caption{Generic hyper-parameters for all tasks.}
\label{table:generic-hyper-param}
\end{table}

\begin{table}[H]
\centering
\begin{tabular}{cccccc}
\toprule
Task & max-depth & train-steps & eval-steps & train-num-constants & eval-num-constants\\
                      \midrule
Predecessor           & 4   & 2     & 4     & 10    & 14    \\
Undirected Edge       & 4   & 2     & 2     & 4     & 6     \\
Less than             & 4   & 12    & 12    & 10    & 12    \\
Member                & 4   & 12    & 12    & 5     & 7     \\
Connectedness         & 4   & 4     & 4     & 5     & 5     \\
Son                   & 4   & 4     & 4     & 9     & 10     \\
Grandparent           & 4   & 4     & 4     & 9     & 11     \\
Adjacent to Red       & 4   & 4     & 4     & 7     & 9     \\
Two Children          & 4   & 4     & 5     & 5     & 7     \\
Relatedness           & 4   & 10    & 12    & 8     & 10    \\
Cyclic                & 4   & 4     & 4     & 6     & 7     \\
Graph Coloring        & 4   & 4     & 4     & 8     & 10     \\
%Length                &    &      &      &      &      \\
Even-Odd              & 4   & 6     & 8     & 11     & 15     \\
Even-Succ2            & 4   & 6     & 8     & 11     & 15     \\
Buzz                  & 4   & 8     & 10     & 11     & 16     \\
%Fizz                  &    &      &      &      &      \\
\bottomrule
\end{tabular}
\caption{Specialized hyper-parameters for each \zj{ILP} task.}
\label{table:spe-hyper-param}
\end{table}

\zj{In multi-task GQA, we use used the same generic hyperparameters as given in Table~\ref{table:generic-hyper-param} except that we disallow recursivity and decreased the learning rate (lr) to $0.001$ for background embeddings and rule learning rate (lr-rules) to $0.01$ for intensional embeddings. Other specific hyperparametes are given in Table \ref{table:MTGQA-hyper-param}.}
\begin{table}[H]
\centering
\begin{tabular}{cc}
\toprule
\textbf{Hyperparameter} & \textbf{Value}  \\
\midrule
Max depth & 3 \\
Train-steps & 4 \\
Embedding-dim & 30 \\
Train-iterations $n_T$ & 3000 \\
Train-num-positive-instances $n_p$ & 5\\
Train-num-random-instances $n_r$ & 5\\
\bottomrule
\end{tabular}
\caption{Multi-task GQA hyperparameters.}
\label{table:MTGQA-hyper-param}
\end{table}
\TODO{Replaced Train-num-round by Train-num-iterations ok?, called n_T}

In reinforcement learning, we used the same generic hyperparameters as given in Table~\ref{table:generic-hyper-param} except that we don't use recursivity and decreased the learning rate to 0.005.
Additionally, hyperparameters are given in Table~\ref{table:rl-hyper-param}.

\begin{table}[H]
\centering
\begin{tabular}{cc}
\toprule
\textbf{Hyperparameter} & \textbf{Value}  \\
\midrule
Max depth & 6\\
Train-steps & 6 \\
Train-num-constants & 5 \\
Temperature of softmax policy & 0.01 \\
GAE $\lambda$ & 0.9 \\
$\gamma$ & 0.99\\
Trajectory per update & 5\\
PPO $\epsilon$-clipping & 0.2\\
GRU hidden neurons (critic) & 64 \\
\bottomrule
\end{tabular}
\caption{Reinforcement learning hyperparameters.}
\label{table:rl-hyper-param}
\end{table}

\subsection{Some promises and limitations of our Model}\label{sec:promise}

\paragraph{Expressivity Limitations}

As made explicit in Theorem\ref{thm:expressivity}, our model corresponds to rule-induction with function-free definite Horn clause. Of course, the number of layers, affects the actual expressivity of each model; 
in the non-recursive case, this parameter is sufficient to reach all the function-free recursive-free definite Horn clause.
In the recursive case, the number of instantiated rules from the proto-rule set (by default set at $1$) also affects the expressivity; this may be seen in the task \texttt{Length} in ILP, where two rules should be initiated per template set to widen sufficient hypothesis space.

\begin{figure}[h]
    \centering
        \includegraphics[width=0.8\columnwidth]{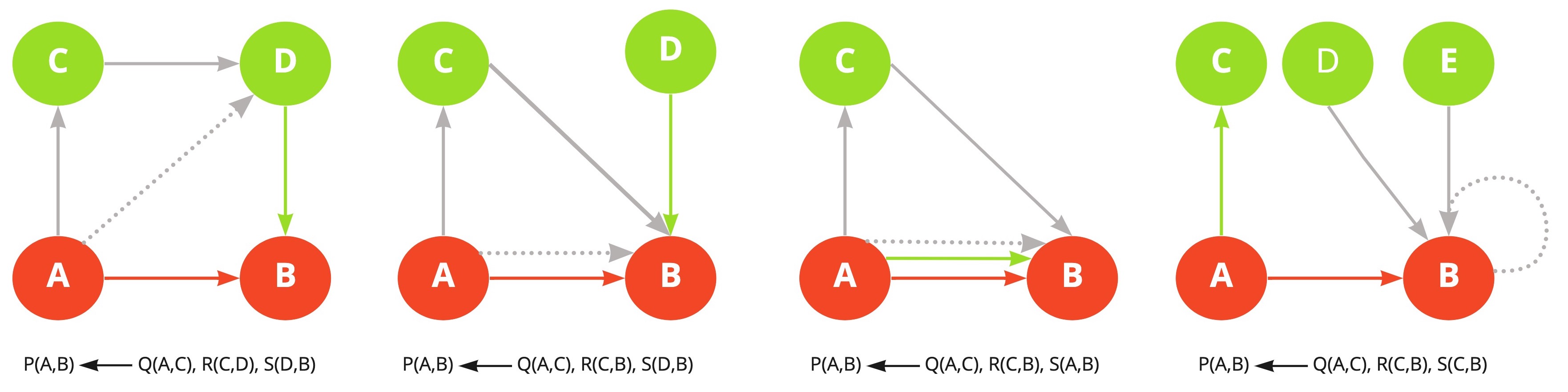} 
    \caption{Reduction of Meta-rules with 3 body atoms}
    \label{fig:3bodyreduction}
\end{figure}

\begin{figure}[h]
    \centering
        \includegraphics[width=0.5\columnwidth]{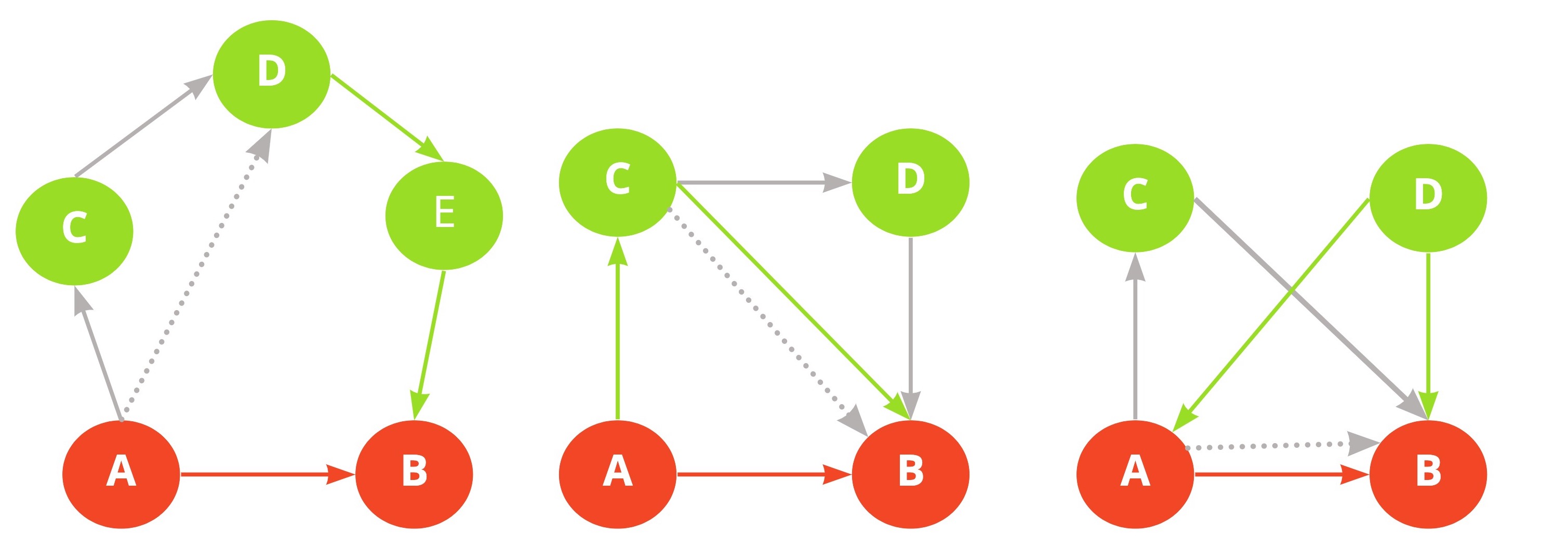} 
    \caption{Reduction of Meta-rules with 4 body atoms}
      \includegraphics[width=0.1\columnwidth]{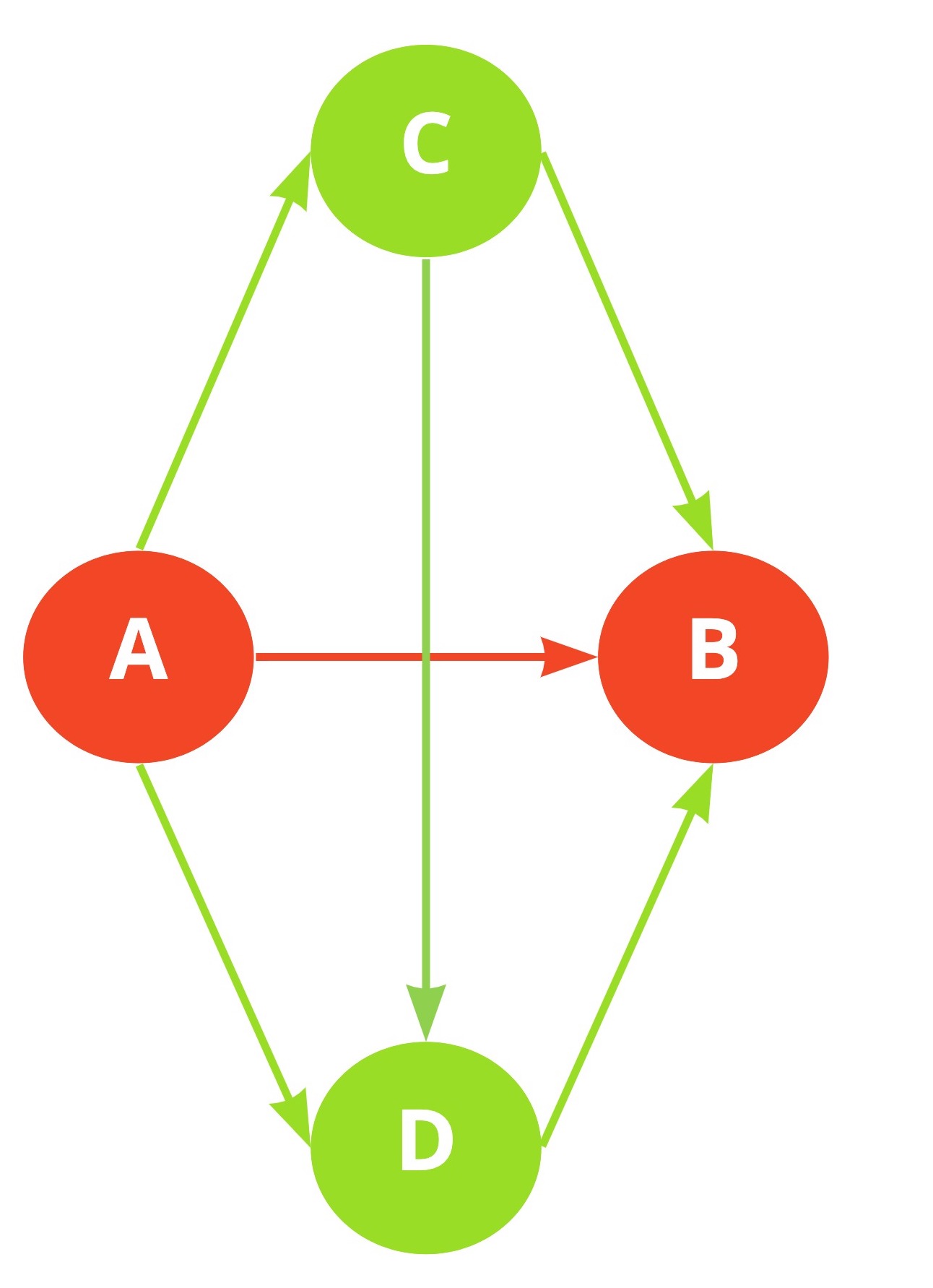} 
    \caption{Irreducible Meta-Rule with 5 body atoms}
    \label{fig:4bodyreduction}
\end{figure}

The expressivity of our model could be extended by different ways, more or less computationally-expensive, and therefore more or less judicious.  The points below would be object of further investigations and experiments; for this reason, we only share in this paper some succinct comments on different tracks to gain expressivity:
\begin{itemize}
\item[+] Enabling negations.
\item[+] Enabling functions.
    \item[+] Enabling zero-ary predicates: 
    We could easily extend our result to include $0$ ary predicates.
   \begin{equation}
       \mathtt{Claim}: \text{By adding the zero-ary predicate } \mathfrak{Z} \text{ to } \PR_0, \text{ the hypothesis space reaches the fragment } \mathcal{F}_{\Predicates, \leq 2}^{\leq 2}
         \end{equation} 
     where $\mathfrak{Z}:\:  P() \leftarrow   \overline{Q}(A, B) $.
    \item[+] Enabling more body atoms:\footnote{For instance the language bias present in \cite{galarraga_amie_2013} is narrowing the space to connected, two-connected and duplicate-free Horn clauses $\mathcal{U}$. On the one hand, $\mathcal{U}$ is a smaller fragment than $\mathcal{F}$ which our model is reaching. However, they consider clauses with $N$ body atoms ($N$ hypothetically small in their experiments), which could enable greater expressivity. }
     \begin{equation} \mathtt{Claim}: \text{Enabling 3 or 4 body atoms would result in the same hypothesis space.}
     \end{equation} 
    This can be proven by reducing clauses with $3$ resp. $4$ body atoms to $2$ resp. $3$ body atoms. Instead of a formal proof, let us gives a visual illustration of these reductions\footnote{We considered clauses upon permutation and symmetries, as usual; we omitted clauses having two body atoms with the same variables, as they can be trivially reduced.}, in Figure \ref{fig:3bodyreduction} resp. Figure \ref{fig:4bodyreduction}.
    Red arrows denote head predicates; full grey arrows indicates two arrows which can be reduced into the dotted grey arrow to lower the number of body atoms by introducing an auxiliary predicate.

   In contrast, as illustrated Figure \ref{fig:4bodyreduction} some rules with 5 body clauses are not reducible to 2 body clauses, such as:
      \begin{equation}
   P(A, B) \leftarrow Q(A, C), R(A, D), S(B, C), T(B, D),U(C, D).   
   \end{equation}
  As mentioned in \citet{cropper_logical_2020},  $\mathcal F_{\leq 5}^{\lbrace 1,2 \rbrace}$ is therefore not D-reducible to  $\mathcal F_{\leq 2}^{\lbrace 1,2 \rbrace}$.
    \pw{However, a}llowing higher number of clauses (such as 5) may drastically increase the computational cost.
    
    \item[+] Enabling higher-arity.\\
    For instance, considering arity $3$ predicates, while keeping two body clauses, could enable to reach $\mathcal{C}_{\leq 5}^{\lbrace 1, 2 \rbrace }$:
   \begin{equation}
       \mathtt{Claim}: \mathcal{F}_{\leq 2}^{\lbrace 1, 2, 3 \rbrace } [\Predicates ] \supset  \mathcal{F}_{\leq 5}^{\lbrace 1, 2 \rbrace }
         \end{equation}

\end{itemize}

Further comparative experiments could be led in the future, to test different minimal or small meta-rules or proto-rules set, and  investigate some judicious balance of minimalism/redundancy and expressivity/efficiency in diverse tasks.

\paragraph{Promises}
%While LRI can obtain good results as indicated in Table~\ref{table:acc-result all} if provided with meta-rules customized for each task, the performance quickly degrades when provided a set of more generic meta-rules (Table~\ref{table:LRI_rule_union_res}).

In contrast to most of previous ILP or differentiable ILP works, we hypothesize that our model could be particularly suited for continual learning in semantically-richer domains; for instance, in RL scenarios where an interpretable logic-oriented higher level policy seems pertinent (e.g., autonomous driving). Here some arguments to support this belief:
\begin{itemize}
    \item[+] In contrast to previous works, our model is independent of the number of predicates, which may vary between training and testing.
    \item[+] As our experiments in Visual Genome suggest, our model can be boostrapped by semantic or visual priors, to initialise the predicates.
    \item[+] In semantically-rich domains, the complexity of our model would be more advantageous than other ILP works. Semantically rich domains are characterized by a high number of initial predicates, while enabling a much lower embedding dimension $d$ (as these predicates are highly interdependent), $d<<\mid \Predicates_0\mid$;
such inequality implies a lower complexity for our model than common methods which have to learn one coefficient per predicate.
    \item[+] The embedding-based approach enables to reason by \emph{analogy}, and possibly quickly generalise to new predicates.
For instance, if the model has learned the rule $OverTake()\leftarrow \lnot (P(X) \land OnNextLane(X))$, and the embedding $\theta_{P}$ is learned close to $\theta_{Car}$, the rule would generalise to include Truck, assuming  $\theta_{Car}\sim  \theta_{Truck}$, despite having never seen the new predicate Truck during training.

\end{itemize}

\end{document}